\documentclass[journal]{IEEEtran}

\usepackage{times}
\usepackage{epsfig}
\usepackage{graphicx}
\usepackage{amsmath,bm}
\usepackage{amssymb}
\usepackage{multirow}
\usepackage{subfigure}
\usepackage{lettrine}
\usepackage{setspace}
\usepackage{algorithmicx}
\usepackage[square, comma, sort&compress, numbers]{natbib}
\usepackage[ruled,vlined]{algorithm2e}

\ifCLASSINFOpdf
\else
\fi
\makeatletter
\def\hlinew#1{%
  \noalign{\ifnum0=`}\fi\hrule \@height #1 \futurelet
   \reserved@a\@xhline}
\makeatother

\begin{document}
\title{Deep Learning Driven Visual Path Prediction from a Single Image}

\author{Siyu~Huang,
        Xi~Li*,
	   Zhongfei~Zhang,
	  Zhouzhou~He,
		Fei~Wu,
		Wei~Liu,
		Jinhui~Tang,
      and Yueting~Zhuang
  
\thanks{S. Huang is with the College
of Information Science \& Electronic Engineering, Zhejiang University, Hangzhou 310027,
China (e-mail: siyuhuang@zju.edu.cn).}
\thanks{X. Li* (corresponding author), F. Wu, and Y. Zhuang are with the College of Computer Science and Technology, Zhejiang University, Hangzhou 310027, China (email:
xilizju@zju.edu.cn, wufei,yzhuang@cs.zju.edu.cn).
}
\thanks{Z. Zhang is with the Department of Information Science and Electronic
Engineering, Zhejiang University, Hangzhou 310027, China, and also with
the Computer Science Department, Watson School, The State University of
New York Binghamton University, Binghamton, NY 13902 USA (e-mail:
zhongfei@zju.edu.cn).}
\thanks{Z. He is with the College
of Information Science \& Electronic Engineering, Zhejiang University, Hangzhou 310027,
China (e-mail: zhouzhouhe@zju.edu.cn). }
\thanks{W. Liu is with the Didi Research, Didi Kuaidi, Beijing 100085, China (e-mail: wliu@ee.columbia.edu).}
\thanks{J. Tang is with the School of Computer Science and Engineering,
Nanjing University of Science and Technology, Nanjing 210094, China
(e-mail: jinhuitang@njust.edu.cn).}
}


\maketitle
\begin{abstract}
Capabilities of inference and prediction are significant components of visual systems. In this paper, we address an important and challenging task of them: visual path prediction. Its goal is to infer the future path for a visual object in a static scene. This task is complicated as it needs high-level semantic understandings of both the scenes and motion patterns underlying video sequences. In practice, cluttered situations have also raised higher demands on the effectiveness and robustness of the considered models. Motivated by these observations, we propose a deep learning framework which simultaneously performs deep feature learning for visual representation in conjunction with spatio-temporal context modeling. After that, we propose a unified path planning scheme to make accurate future path prediction based on the analytic results of the context models. The highly effective visual representation and deep context models ensure that our framework makes a deep semantic understanding of the scene and motion pattern, consequently improving the performance of the visual path prediction task. In order to comprehensively evaluate the model's performance on the visual path prediction task, we construct two large benchmark datasets from the adaptation of video tracking datasets. The qualitative and quantitative experimental results show that our approach outperforms the existing approaches and owns a better generalization capability.
\end{abstract}


\IEEEpeerreviewmaketitle

\section{Introduction}
\begin{figure}[t]
\centering
\subfigure[Original Image]{
\begin{minipage}[b]{0.46\linewidth}
\includegraphics[width=1\linewidth]{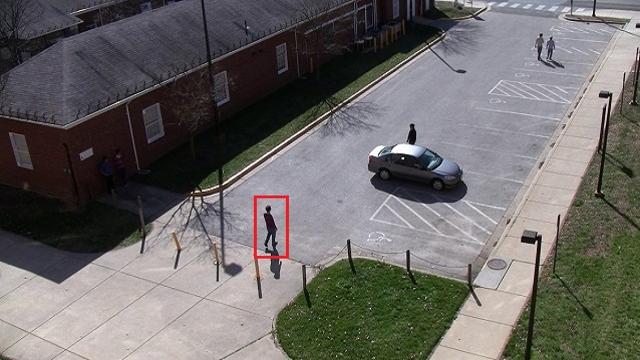}
\end{minipage}
}
\subfigure[Reward Map]{
\begin{minipage}[b]{0.46\linewidth}
\includegraphics[width=1\linewidth]{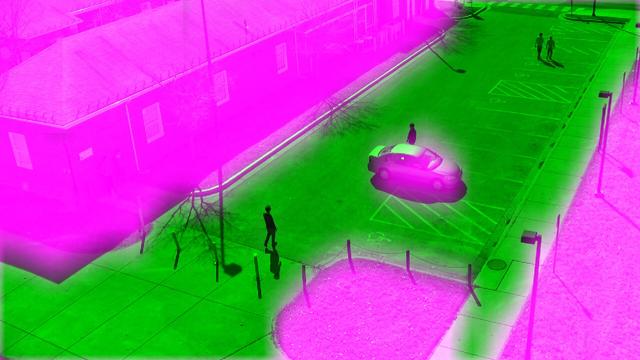}
\end{minipage}
}
\subfigure[Estimated Orientation]{
\begin{minipage}[b]{0.46\linewidth}
\includegraphics[width=1\linewidth]{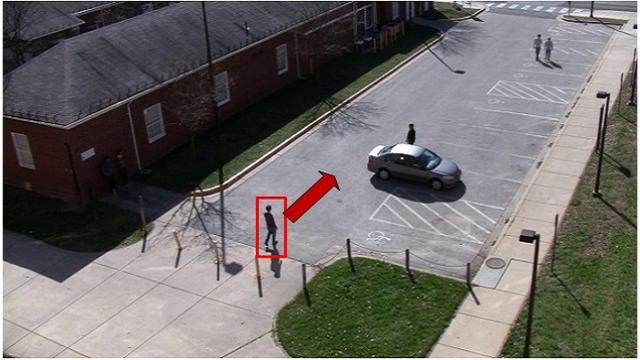}
\end{minipage}
}
\subfigure[Predictions]{
\begin{minipage}[b]{0.46\linewidth}
\includegraphics[width=1\linewidth]{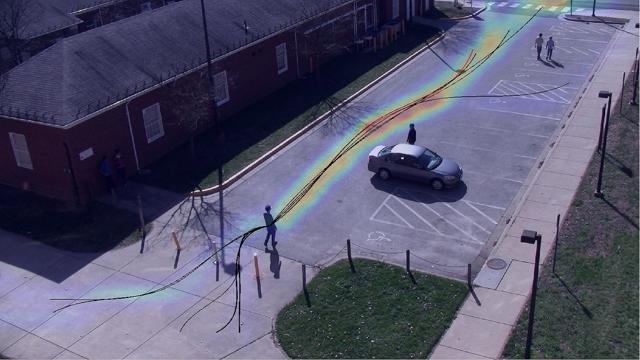}
\end{minipage}
}
\caption{Illustration of our approach. Image (a) shows a man in the parking lots. The goal of visual path prediction is to infer the possible paths for him in the future. In this paper, we first generate a reward map (b) representing regions the man can reach in the future (green). Then, we estimate his facing orientation (c). Finally, we incorporate the results of (b) and (c) to plan the most likely paths as shown in (d), where the red line and the black lines respectively show our top-1 and top-10 predictions.}
\label{fig:1}
\end{figure}

\lettrine[lines=2]{\textbf{I}}{nference} and prediction are significant capabilities of intelligent visual systems \cite{hawkins2007intelligence} such that they have been popular topics in computer vision community during recent years. As part of visual inference and prediction, we address the visual path prediction problem, with the goal inferring the most possible future path for an object in a static scene image. For instance, given a single static image like Fig. \ref{fig:1}(a), we humans can easily recognize the objects inside it, and tell others which ones are active --- persons and car will move, but grass and house will remain still. Furthermore, for the active objects, we will naturally infer their intentions and future motions. Taking the man at bottom left with red bounding box for an example, he is most likely to walk straight, meanwhile, bypassing the car which appears to be an obstacle for him. The aforementioned visual inference process is illustrated as the red path in Fig. \ref{fig:1}(d). As a matter of fact, these predictions are naturally driven by a human visual system and supported by the prior knowledge stored in it.

In this work, we aim to automatically learn this prior knowledge from a diverse set of videos, and further infer the possible future motions of objects. The prior knowledge here includes both the scene structure and motion patterns underlying the frame sequences. More specifically, they can be respectively associated with the contextual properties of the scene structure from the spatio-temporal perspectives. Therefore, the key way of solving the visual path prediction task is modeling the spatial and temporal context, followed by a certain inference algorithm to predict the future path. Such a task is very challenging because it not only needs deep semantic understanding of videos, but also is often confronted with very complicated and diverse situations. For instance, just a single scene in this task may contain various kinds of appearance which are easy to confuse with each other. To address this dilemma, the visual representation is typically required to be semantic and highly discriminative. On the other hand, the scenes and objects are usually diverse that cover a large amount of cases. It is required for the context model and visual representation to possess good enough generalization ability for the adaptation to complex scenarios.

In recent years, topics of visual inference and prediction are widely studied by computer vision reseachers, and there has been some work referring to the visual path prediction task. Earlier work \cite{liu2008sift,yuen2010data} focuses on the matching-based approaches. For instance, Yuen et al. \cite{yuen2010data} explore scene modeling by searching straightforwardly in image space with keypoint matching techniques using descriptors like GIST and dense SIFT. In general, these matching-based methods rely on large amount of data and do not really understand the scene. In the test phase, they have to compare with all the alternative samples, leading to high computation cost. In contrast, more recent work has poured attention into the learning-based approaches \cite{kitani2012activity,walker2014patch}. The key concept is learning context model to capture the structure relationships between the scene and specific objects, followed by learning robust temporal models like IOC \cite{kitani2012activity} for inference. The learning-based approaches seek to establish inductive models to understand the scene in depth, which results in the state-of-the-art performance in the visual path prediction task. 

While in practice, the complex and cluttered situations (e.g., a crowd of cars and people moving at the crossroads) in this task have raised higher demands on the effectiveness and robustness of our models. In general, the conventional visual representations are based on handcrafted features, which are often much restrictive in complex visual scenes and thus cannot provide abundant semantic information about the visual content. Besides, the context model built in the aforementioned approaches is relatively simple and shallow, which leads to the inability of modeling the intrinsic contextual interactions among objects as well as their associated scene structures. For instance, Walker et al. \cite{walker2014patch} build their context model by straightforwardly counting the votes from training data. Such a practice is hard to effectively model the contextual information. 

Motivated by these observations, in this paper we propose a unified deep learning framework for visual path prediction, which simultaneously performs deep feature learning for visual representation
in conjunction with spatio-temporal context modeling. After that, a unified path planning scheme is proposed to make accurate future path prediction based on the analytic results of the context models. Compared with the conventional approaches to visual path prediction, the visual representation employed in our framework is highly effective because it has a better discrimination and generalization capability. Meanwhile, our deep context models can be better adapted to the complex scenarios. These improvements ensure that our framework can make a deep semantic understanding about the scene and motion pattern, consequently improving the performance in the visual path prediction.  
 
The key contributions of our paper are summarized as follows:
\begin{enumerate}
\item 
We present a novel deep learning framework based on the CNNs for visual path prediction task. To the best of our knowledge, it is the first work to leverage a deep learning approach in this task. Our framework models both of the scene structure information and motion patterns. The abstraction of visual representation and the learning of context models are accomplished in a unified framework. It largely improves the scene understanding capability compared with the previous approaches.
\item 
We propose a unified path planning scheme to infer the future paths on the basis of the analytic results returned by our context models. In this scheme, the problem of future path inference is equivalently converted into an optimization problem, which can be solved in an efficient way.
\item
We construct two benchmark datasets for visual path prediction from the adaptation of two large video tracking datasets \cite{oh2011large, weisbrich2014kit}. The adapted datasets are much larger than those used in the previous work and cover a diverse set of scenes and objects. They can be used for comprehensively evaluating the performance of a visual path prediction model. They will be publicly available on our homepage.     
\end{enumerate}

\section{Related Work}

In general, the methods for visual path prediction contain two components: (1) Understanding the scene and motion pattern of the video sequences. (2) Inferring the future path based on information obtained by step (1). This section will review the representative methods of these two steps respectively.

\noindent
{\bf Understanding the scene and motion pattern:} Scene understanding is a prerequisite to many high level tasks for intelligent systems operating in real world environments. In the past ten years, researchers have made great efforts for understanding the static image scene at a deeper level, including several typical topics like scene classification \cite{li2012objects,yao2012describing,yu2013pairwise,juneja2013blocks,karpathy2014large}, semantic segmentation and labeling \cite{silberman2011indoor,farabet2013learning,girshick2014rich,zhang2014probabilistic,li2014geodesic}, depth estimation \cite{saxena20083,liu2010single,lin2013absolute}, etc. What these approaches have in common is that they learn and model the scene structure prior to recover different aspects of a scene. Accordingly, Yao et al. \cite{yao2012describing} propose a holistic structure prediction model based on CRF to jointly solve several scene understanding problems. 

Except for modeling scenes in static images and inferring knowledge at the current state, an intelligent visual system is supposed to be able to infer what will happen in the near future. In more recent years, many researchers have paid attention to modeling the motion pattern in video sequences for temporal aspect of recognition and prediction. For instance, recognition and forecasting of human action \cite{kitani2012activity,koppula2013anticipating,nascimento2013activity,lan2014hierarchical,wang2014action,wang2014latent}, event \cite{yuen2010data,duan2012visual,merler2012semantic,jiang2013high} and scene transition \cite{fouhey2014predicting,walker2014patch} have caught lots of interest. For dynamic scene understanding, the key is to model the structure relationships among different frames. As well, techniques of static scene understanding play a significant role in it.
 
\noindent
{\bf Path inference:} 
Methods for path inference can be generally classified into two categories: the matching-based methods \cite{liu2008sift,yuen2010data,keller2011will} and the learning-based methods \cite{kitani2012activity,walker2014patch}. The matching-based methods simply retrieve the information from databases to the queries without building an inference model. For instance, Liu et al. \cite{liu2008sift} propose a method by matching a query image to a large amount of video data and warping the ground truth paths from the nearest neighbour videos to the static query image with SIFT Flow. Instead of the warping process, Yuen et al. \cite{yuen2010data} build localized motion maps as probability distributions after merging votes from several nearest neighbors. These matching-based approaches rely on the richness of the databases. 

On the other hand, the learning-based methods learn temporal inference models to capture the spatio-temporal variation of scenes and objects. Temporal models such as Markov Logic Networks \cite{tran2008event}, IOC \cite{kitani2012activity}, CRF \cite{fouhey2014predicting}, ATCRF \cite{koppula2013anticipating} and EDD \cite{lampert2015predicting} are often employed. These models help infer the future of individual objects. Further work has taken into consideration the relationships between objects and scenes. Kitani et al. \cite{kitani2012activity} detect the physical scene features based on semantic scene labeling techniques \cite{munoz2010stacked,munoz2012co}, and then, fuse them into the reward function of IOC. Walker et al. \cite{walker2014patch} build a temporal model based on the effective mid-level patches \cite{singh2012unsupervised}. They learn patch-to-patch transition matrix, which serves as the temporal model, and learn a context model for the interaction between mid-level patch and the scene. These approaches draw on the strength of scene semantic understanding in depth, and successfully advance the overall performance for visual path prediction task. 

\noindent
{\bf Convolutional Neural Networks:} The proposed framework in this paper is built upon the convolutional neural networks (CNNs). The CNNs are a popular and leading visual representation technique, for they are able to learn powerful and interpretable visual representations \cite{karpathy2014large}. The CNNs have given the state-of-the-art performance on various computer vision tasks \cite{krizhevsky2012imagenet,ciresan2012multi,farabet2013learning,karpathy2014large,girshick2014rich}. In recent years, some work has combined CNNs with temporal modeling. Donahue et al. \cite{donahue2015long} use CNNs and LSTM \cite{hochreiter1997long} for visual recognition and description tasks. On the perspective of temporal prediction, Walker et al. \cite{walker2015dense} employ the same architecture of \cite{donahue2015long} to predict long term motion of pixels in terms of optical flow. 

\section{Our Approach}
\label{section:approach}

\subsection{Problem Formulation}

\begin{figure}[t]
\begin{center}
   \includegraphics[width=1\linewidth]{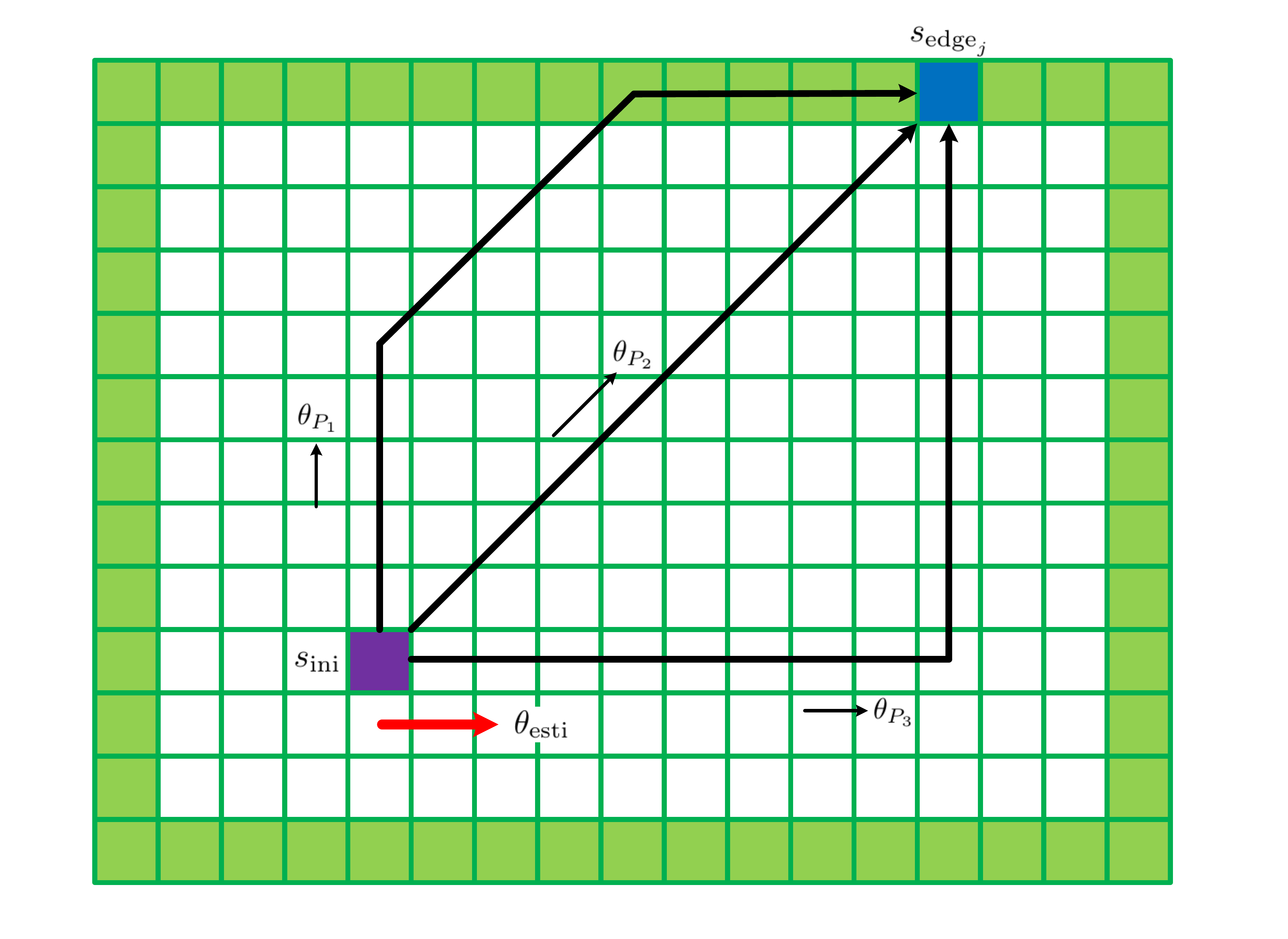}
\end{center}
\caption{A simple illustration of the visual path prediction problem. Between the object location $s_{\mathrm{ini}}$ and the $j$-th edge point $s_{\mathrm{edge}_j}$, we desire to plan a path $P$ which has the lower spatial matching costs on the cost map, meanwhile, the smaller angular difference between its initial moving direction $\theta_P$ and the estimated direction $\theta_{\mathrm{esti}}$.
} 
\label{fig:planningscheme}
\end{figure}

In this work, we aim to build a framework to automatically solve the visual path prediction problem. Given a static scene image $I$ and the bounding box $B=(b_1,b_2,w,h)$ of an object in $I$, the goal is to infer the most possible path $P=(s_1,s_2,s_3,\dots,s_n)$ of the object in the future. Here, $b_1,b_2,w,h$ are respectively the top left coordinate, the width, and the height of $B$. And $s=(x,y)$ represents the coordinate of a position, such that $P$ consists of a sequence of adjacent positions. Fig. \ref{fig:planningscheme} gives an illustration of the problem. We formalize the original scene into a grid graph such that each grid corresponds to a specific position $s_i$ of the scene. Between the object location $s_{\mathrm{ini}}$ (the center of $B$) and a certain edge location $s_{\mathrm{edge}_j}$, there are a large amount of alternative paths. The question is how to select such an appropriate path $P$ from the very large path space $\mathbb{P}$? We convert the original problem into an optimization problem of planning a path $P$ with the lowest cost $\mathcal{C}$:
\begin{align}
\label{eq:generaloptimization}
& \min_{P} ~ \mathcal{C}(P), \\
& ~\text{s.t.}~~  P \in \mathbb{P}. \notag 
\end{align}

Then, the issue is how to formulate the cost $\mathcal{C}$ of a path $P$. Intuitively, if there are more obstacles on a path, the associated cost of it ought to be higher:
\begin{equation}
\label{eq:spatialcost}
\mathcal{C}_{\mathrm{S}} (P) =  \sum_{s_{i} \in P} \bm{R}_{\mathrm{cost}}(s_{i}).
\end{equation}

\noindent
$\bm{R}_{\mathrm{cost}}$ is a cost map of the scene representing the cost of each coordinate position $s_i$. Therefore, we need to discover which regions of the scene the object can reach. Such a structure relationship between the object and the scene is referred to as ``spatial context matching''; thus $\mathcal{C}_{\mathrm{S}}$ is referred to as the ``spatial matching cost'' in this paper. We build a deep context model called \textbf{Spatial Matching Network} to learn the spatial contextual information from the video sequences in the training phase. In the testing phase, Spatial Matching Network generates a cost map $\bm{R}_{\mathrm{cost}}$ according to a testing scene image. 

On the other hand, the object's current moving direction also crucially influences the path selection. Hence, paths which are consistent with the object's current moving direction $\theta_{\mathrm{GT}}$ should have lower costs:
\begin{equation}
\label{eq:orientationcost}
\mathcal{C}_{\mathrm{O}}(P)=\mathcal{D}(\theta_P,\theta_{\mathrm{GT}}).
\end{equation}

\noindent
Here, $\mathcal{C}_{\mathrm{O}}$ is called as ``orientation cost'' in this paper. $\theta_P$ is the initial moving direction of $P$, and $\mathcal{D}(\theta_1,\theta_2)$ represents the angular difference between two angles $\theta_1$ and $\theta_2$. For the sake of motion orientation modeling, we build another deep context model called \textbf{Orientation Network} to learn the temporal contextual information underlied in video sequences. In the testing phase, Orientation Network estimates an object's facing orientation $\theta_{\mathrm{esti}}$ as $\theta_{\mathrm{GT}}$ from the single object image.

The above two types of costs --- the spatial matching cost $\mathcal{C}_{\mathrm{S}}$ and the orientation cost $\mathcal{C}_{\mathrm{O}}$ adequately help us semantically understand the scene and make a decision about the future path. As shown in Fig. \ref{fig:planningscheme}, suppose the three paths $P_1,P_2,P_3$ have the same average accumulated costs on cost map $\bm{R}_{\mathrm{cost}}$. Which one is optimal? $P_3$ wins out because its initial direction $\theta_{P_3}$ is closer to $\theta_{\mathrm{esti}}$. Therefore, the path cost $\mathcal{C}$ is written as
\begin{equation}
\label{eq:costplus}
\mathcal{C}(P)=\mathcal{C}_{\mathrm{S}} (P) + \varepsilon \mathcal{C}_{\mathrm{O}}(P),
\end{equation}

\noindent
where $\varepsilon$ is a trade-off coefficient between the two types of costs. Finally, substituting $\mathcal{C}(P)$ into the optimization problem \eqref{eq:generaloptimization}, we propose a unified path planning scheme to solve it in an easy and efficient way.

\begin{figure*}[t]
\begin{center}
   \includegraphics[width=1\linewidth]{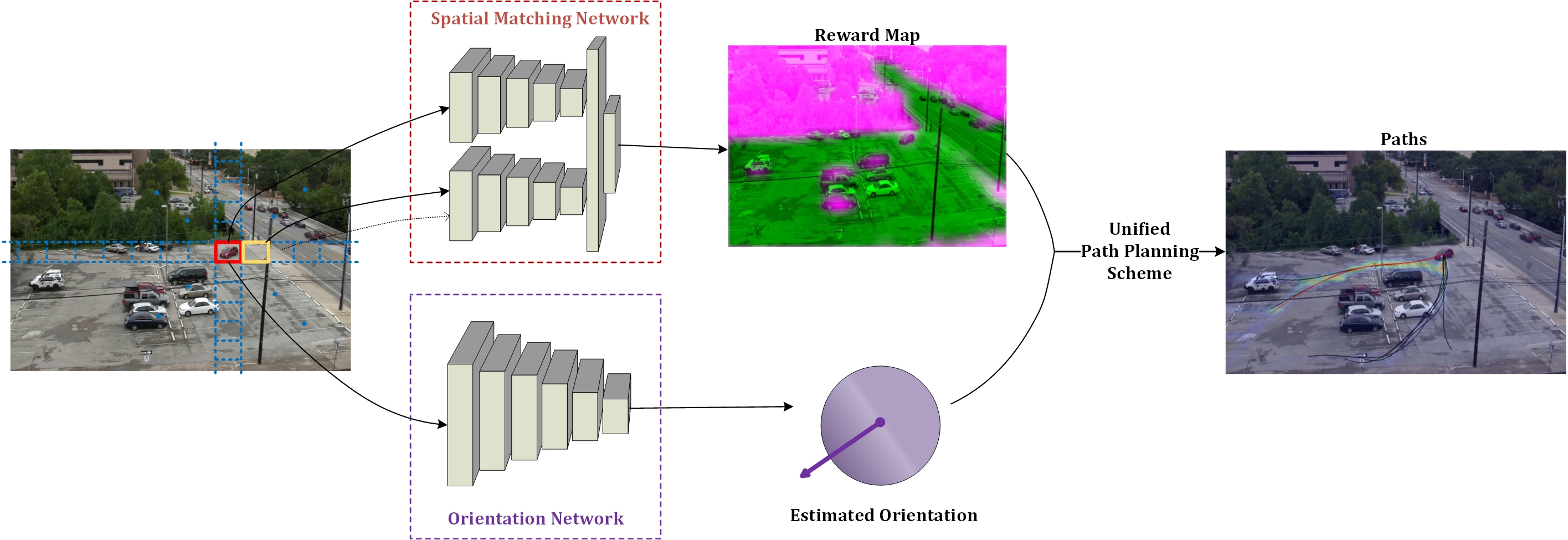}
\end{center}
\caption{The overview of our framework. Spatial Matching Network and Orientation Network are two CNNs, which respectively model the spatial and temporal contexts. We repeatedly input images of the object and local environment patches into Spatial Matching Network to generate the reward map of the scene. Intuitively, it helps us decide whether the object could reach certain areas of the scene. Orientation Network estimates the object's facing orientation, which indicates the object's preferred moving direction in the future. Then we incorporate this analysis and infer the most likely future paths with a unified path planning scheme. }
\label{fig:framework}
\end{figure*}

Fig. \ref{fig:framework} shows our general framework in the testing phase. The far left of the figure is the input of visual path prediction problem, containing a scene image of parking lots and a bounding box of a car. We employ two CNNs to semantically analyze different aspects of the scene and the object. The first CNN, which we call Spatial Matching Network, generates a reward map $\bm{R}_{\mathrm{reward}}$ representing the reward of every pixel on the scene image. The larger reward means the higher probability the car will reach that pixel position in the future. The reward map $\bm{R}_{\mathrm{reward}}$ is then converted into a cost map $\bm{R}_{\mathrm{cost}}$ for the subsequent path planning. The second CNN, which we call Orientation Network, outputs an estimated facing orientation $\theta_{\mathrm{esti}}$ of the car. And then, based on the analytic results $\bm{R}_{\mathrm{cost}}$ and $\theta_{\mathrm{esti}}$, we infer the most possible future paths of the car by solving the optimization problem \eqref{eq:generaloptimization}. 

In such a framework, there are still some important problems to solve in what follows: For the two networks, how do we learn the contextual information from video sequences, and, what are the appropriate architectures of them? How do we efficiently solve the optimization problem \eqref{eq:generaloptimization}? We will discuss these issues in the following subsections.

\subsection{Spatial Matching Network}

\begin{figure}[t]
\begin{center}
   \includegraphics[width=1\linewidth]{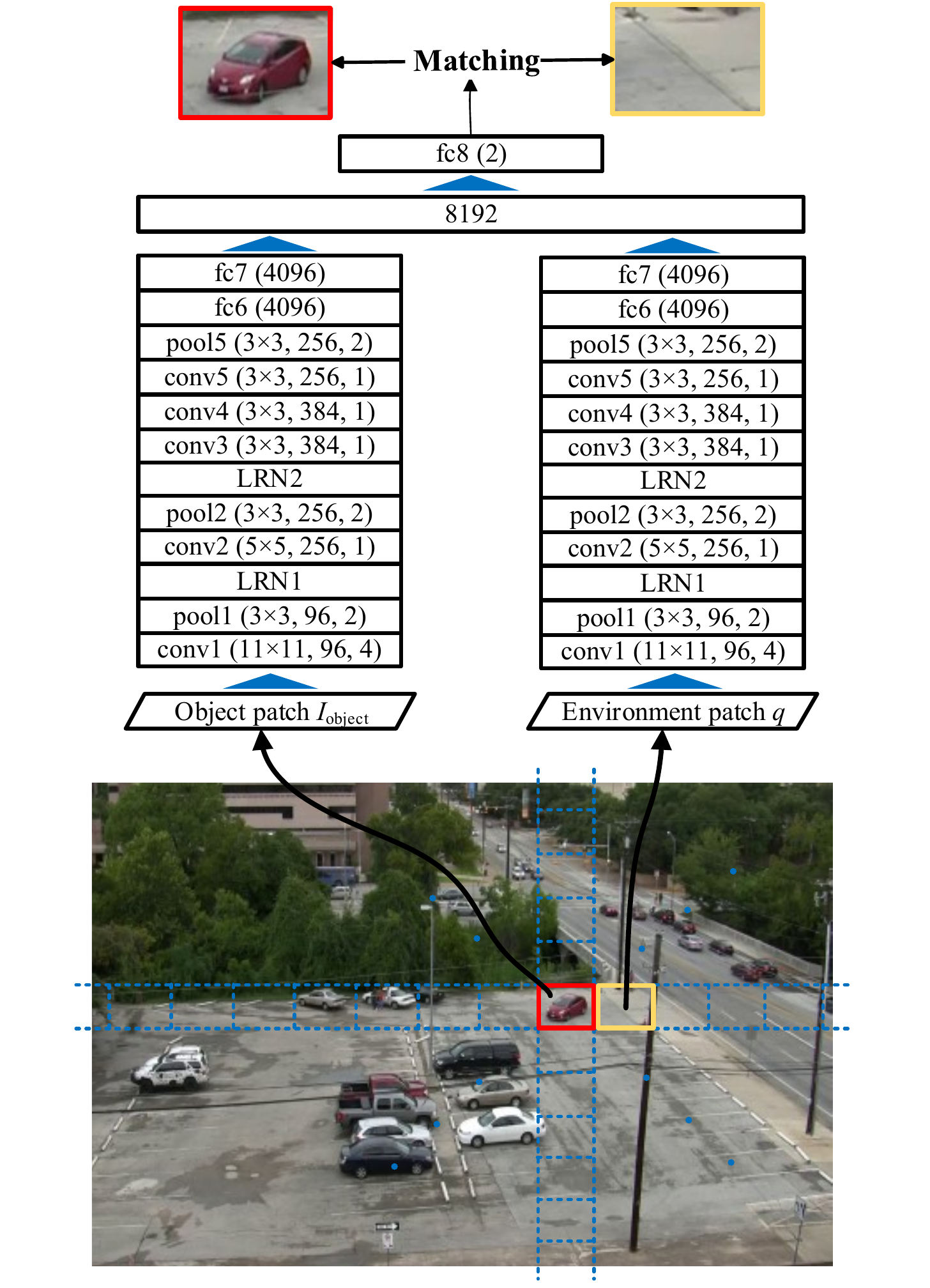}
\end{center}
\caption{Illustration of Spatial Matching Network. The bottom is the scene image. We crop out the object image patch $I_{\mathrm{object}}$ with the bounding box shown in red. We use a sliding window on the entire scene image to crop out the local environment patches, shown as the blue boxes with dotted lines. Each time we input the object patch $I_{\mathrm{object}}$ and an environment patch $q$ into the network. The network outputs the likelihood of spatial context matching between the two patches. In this figure, two inputs are the car and the ground. They are spatial context matching, so label $L_{\mathrm{S}}$ for this sample is set as 1 during training.}
\label{fig:Spatial Matching Network}
\end{figure}

We build Spatial Matching Network to model the interaction relationships between various objects and regions in scenes, namely the spatial context. More intuitively, for example, a pedestrian is more likely to walk on the pavement than climbing over the fence beside it. Here we call the pedestrian and the pavement as spatial context matching, while the pedestrian and the fence are not spatial context matching. As another example, if there is a house in front of a car, the car is supposed to detour the house but not to crash into it. Obviously the car is not spatial context matching with the house in this case. In our framework, such relationships are modeled by Spatial Matching Network. 

Fig. \ref{fig:Spatial Matching Network} illustrates the architecture of Spatial Matching Network. We expect the network to model the relationship between two instances, so it takes two image patches as its input at the same time. One represents the given object and the other is a certain local environment patch obtained by a sliding window on the entire scene image, denoted as the blue boxes with dotted lines shown in Fig \ref{fig:Spatial Matching Network}. The two inputs respectively propagate through two CNNs from conv1 to fc7 and then concatenated into a new fully connected layer fc8. The layers from conv1 to fc7 are similar to the AlexNet \cite{krizhevsky2012imagenet}. Note that the parameters of the two CNNs are different, as their inputs come from two different semantic spaces. We use a softmax layer at the output end of Spatial Matching Network. In the training phase, the label $L_{\mathrm{S}} \in \{0,1\}$ of the network is set as 1 if the two input patches are spatial context matching. Otherwise it is set as 0. In the testing phase, the network outputs the likelihood $r$ of spatial context matching between the object patch $I_\mathrm{object}$ and the local environment patch $q$:
\begin{equation}
r=\mathcal{F}_{\mathrm{S}} ( I_{\mathrm{object}}, q; \psi_{\mathrm{S}} ) .
\end{equation}

\noindent
$I_{\mathrm{object}}$ is obtained according to the object bounding box $B$ on the scene image $I$. $\mathcal{F}_{\mathrm{S}}$ represents the forward propagation in Spatial Matching Network, and $\psi_{\mathrm{S}}$ are its learned parameters. For a scene image $I$, we can crop out the local environment patches $\bm{q}=\{q_{s_1},q_{s_2},\ldots,q_{s_t}\}$ with an overlapped sliding window on $I$, where $s_i=(x_i,y_i)$ is the central position of patch $q_{s_i}$. In this way, we can generate a reward map $\bm{R}_{\mathrm{reward}}$ for an object $I_{\mathrm{object}}$ and a scene image $I$ by repeatedly inputing all the local environment patches $\bm{q}$ with the same object patch $I_{\mathrm{object}}$ into Spatial Matching Network:
\begin{equation}
\bm{R}_{\mathrm{reward}}(s_i)=\mathcal{F}_{\mathrm{S}} ( I_{\mathrm{object}}, q_{s_i}; \psi_{\mathrm{S}} ).
\end{equation}

\noindent
$\bm{R}_{\mathrm{reward}}(s_i) \in [0,1]$ is the reward $r$ for each position $s_i$. The larger value means the higher reward for that position, namely the higher probability the object will reach that position in the future. Visualization of our reward maps generated on different scenes are shown in the middle column of Fig. \ref{fig:qualitative}. 

It is noted that the reward function in the previous work \cite{kitani2012activity,walker2014patch} only models the scene itself. However, different objects may have different relationships with the same region of the scene. So the reward map in our method is built with respect to both of the specific object and the scene appearance, for the purpose of generalization across a diverse set of scenes and objects. 

The reward map $\bm{R}_{\mathrm{reward}}$ can be converted to the cost map $\bm{R}_{\mathrm{cost}}$, such that:
\begin{equation}
\label{rcost}
\bm{R}_{\mathrm{cost}}(s)=\dfrac{1}{1+e^{-\alpha(\bm{R}_{\mathrm{reward}}(s)-\gamma)}},
\end{equation}

\noindent
where $\alpha$ is the tolerance to obstacles. $\gamma$ is fixed to $0.5$, as the scale of $\bm{R}_{\mathrm{reward}}(s)$ is $[0,1]$. Based on this formulation, we can compute the spatial matching cost $\mathcal{C}_{\mathrm{S}}$ of a path $P$ according to Eq. \eqref{eq:spatialcost}. 

\subsection{Orientation Network}

In this subsection, we discuss how to build Orientation Network to learn the temporal context from video sequences in the training phase, and, to estimate an object's facing orientation $\theta_{\mathrm{esti}}$ in the testing phase. Because the scene is assumed to be static in the visual path prediction task, we only focus on modeling the temporal context of the object itself. In other words, we are going to model the time-dependent variation of the object's own state. The state here includes the physical appearance and the spatial position. As the information about physical appearance has been integrated in Spatial Matching Network, we only model the position variation of object itself, namely the relative position of the object at different time. When in the test phase, it is represented as the object's facing orientation with the input of a single image. The temporal context also plays an important role in selecting the future path. For instance, imagine a man walking on the street; he is most likely to walk along his facing orientation. Similarly, any kind of active object follows this rule if there are no other external factors disturbing it. 

\begin{figure}[t]
\begin{center}
   \includegraphics[width=1\linewidth]{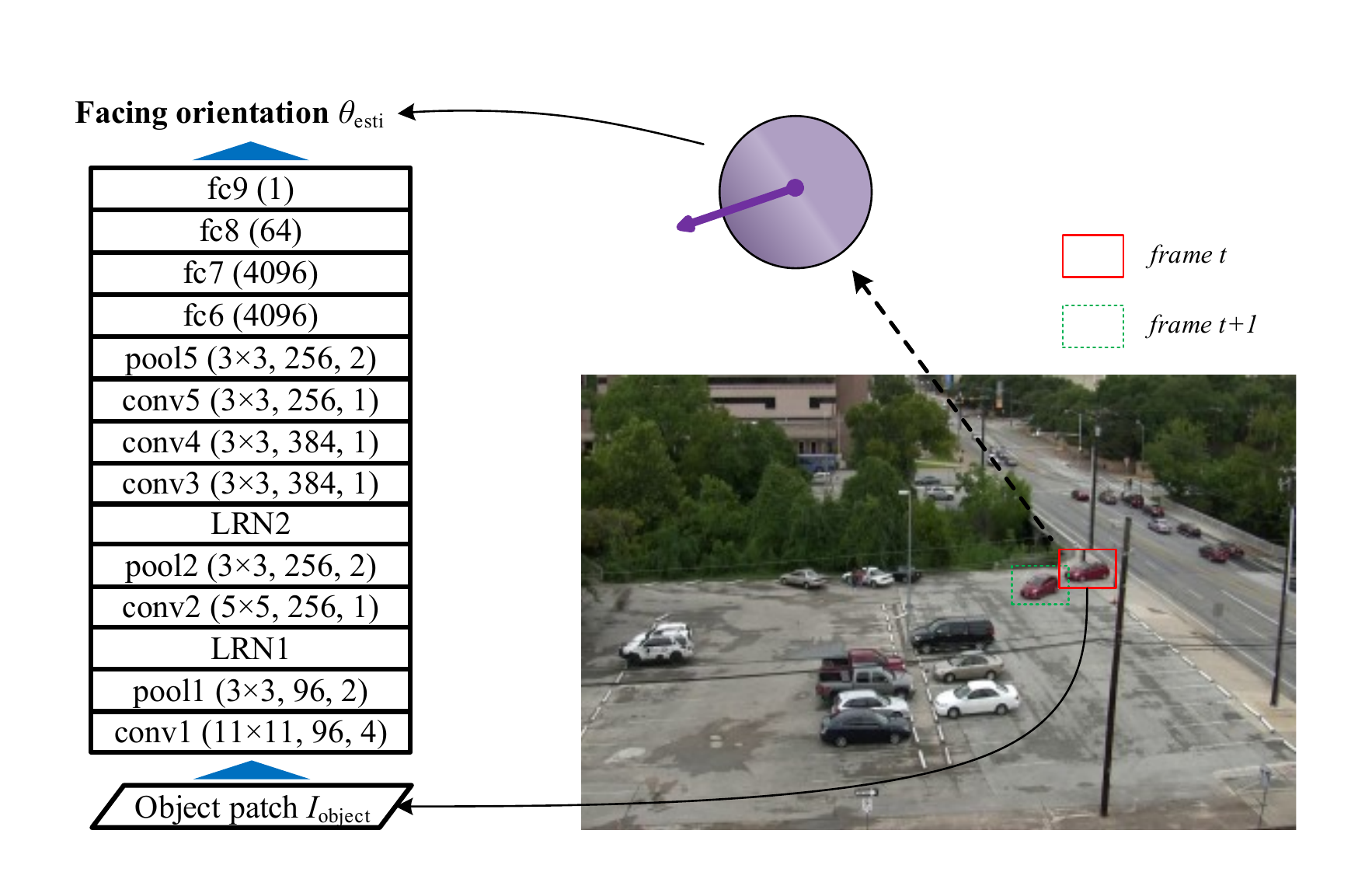}
\end{center}
\caption{Illustration of Orientation Network. What is the facing orientation of the given object? We train Orientation Network to estimate it accurately. The network takes an object image $I_{\mathrm{object}}$ as input. It outputs the estimated facing orientation angle $\theta_{\mathrm{esti}}$ of the object. The architecture from conv1 to fc7 is similar to AlexNet \cite{krizhevsky2012imagenet}. We add fc8 and fc9 to reduce features' dimension, followed by a regression layer. The relative position of the same object between neighbouring frames serves as the ground truth label.}
\label{fig:orientation network}
\end{figure}

Therefore, we build Orientation Network to estimate an object's facing orientation $\theta_{\mathrm{esti}}$. The architecture of Orientation Network is shown in Fig. \ref{fig:orientation network}. We first extract image features using the standard seven-layer architecture similar to AlexNet \cite{krizhevsky2012imagenet} and then embed the features to low-dimensional space with linear mapping. At the output end of Orientation Network, the low-dimensional features are finally regressed into a single value $\theta_{\mathrm{esti}} \in (-\pi,+\pi]$, which represents the estimated angle of the object's facing orientation. Now we decide an appropriate loss function. Intuitively, the orientation estimation can be posed as either classification or regression. Walker et al. \cite{walker2014patch} treat it as classification because the state space of their temporal model is discrete. However, the orientation angle has reasonably high spatial self-correlation. The labels in classification task are often not sufficiently related to each other or even mutually exclusive. Therefore, we are going to use regression as the output of Orientation Network, in view of its correlation between labels. We use Euclidean distance as the regression loss $loss_{\mathrm{o}}$ of Orientation Network:
\begin{equation}
\label{loss}
loss_{\mathrm{o}}=\mathcal{D}{^2}(\theta_{\mathrm{GT}},\theta_{\mathrm{esti}}),
\end{equation}

\noindent
where $\theta_{\mathrm{GT}} \in (-\pi,+\pi]$ is the ground truth angle set as the relative position of the same object between two neighbouring frames, and $\theta_{\mathrm{esti}}$ is the output of Orientation Network. $\mathcal{D}(\theta_1,\theta_2)$ is the angular difference between two angles $\theta_1$ and $\theta_2$:
\begin{equation}
\label{angledifference}
\mathcal{D}(\theta_1,\theta_2)=
\begin{cases}
|\theta_1-\theta_2|  ~~~~  & |\theta_1-\theta_2|\leqslant\pi, \\
2\pi-|\theta_1-\theta_2|  &  |\theta_1 -\theta_2|>\pi.
\end{cases}
\end{equation}

In the testing phase, we can estimate the facing orientation $\theta_{\mathrm{esti}}$ of the input object image $I_{\mathrm{object}}$ by doing forward propagation $\mathcal{F}_{\mathrm{O}}$ in Orientation Network:
\begin{equation}
\theta_{\mathrm{esti}}=\mathcal{F}_{\mathrm{O}} ( I_{\mathrm{object}};\psi_{\mathrm{O}} ) ,
\end{equation}

\noindent
where $\psi_{\mathrm{O}}$ are the learned parameters of Orientation Network.

\subsection{Path Planning}

Up to now, the contextual properties of the scene structure are respectively formalized to be a cost map $\bm{R}_{\mathrm{cost}}$ corresponding to the scene and an estimated facing orientation $\theta_{\mathrm{esti}}$ corresponding to the object. How do we plan the most probable future path for the given object? We propose a unified path planning scheme by efficiently solving the primitive optimization problem \eqref{eq:generaloptimization}. The right part of Fig. \ref{fig:framework} illustrates the function of this scheme. 

The optimization problem \eqref{eq:generaloptimization} aims to find the optimal path $P=(s_1,s_2,\dots,s_n)$ from the path space $\mathbb{P}$, which has the lowest path cost $\mathcal{C}(P)$. By combining Eq. \eqref{eq:spatialcost}, \eqref{eq:orientationcost}, \eqref{eq:costplus} and a few constraints to $\mathbb{P}$, we rewrite problem \eqref{eq:generaloptimization} as:
\begin{align}
\label{optimization}
\min_{P} &   \sum_{s_{i} \in P} \bm{R}_{\mathrm{cost}}(s_{i}) + \varepsilon  \mathcal{D}(\theta_P,\theta_{\mathrm{esti}}), \\
\text{s.t.}~~ & s_{i} ~\text{and}~s_{i+1} ~ \text{are spatially adjacent}, \notag \\
&s_{1}=s_{\mathrm{ini}},  \notag \\
&s_{n}=s_{\mathrm{edge}_j},~j=1,\ldots,m, \notag
\end{align}

\noindent
where the first constraint means that the object can only move to one of its adjacent positions in every step. In our experiments we use eight directions (top, left, bottom, right, top-left, top-right, bottom-left, bottom-right). The second and the third constraints specify the starting and ending positions of paths, where $m$ is the number of edge points. The initial moving direction $\theta_P$ of $P$ is obtained by computing the relative position between the initial position $s_{\mathrm{ini}}$ and a certain position $s_i$ on $P$. In our experiments, the distance $d$ between $s_{\mathrm{ini}}$ and $s_i$ is fixed to the diagonal length of the object bounding box $B$: $d=\lfloor\sqrt{w^2+h^2}\rfloor$, where $\lfloor\cdot\rfloor$ is the rounding floor. $\varepsilon$ is set to 5 as a matter of experience.

\begin{algorithm}[t]
\caption{Visual path planning framework}\label{alg}
\DontPrintSemicolon
\SetKw{Initialize}{Initialize}
\SetKwInOut{Input}{Input}
\SetKwInOut{Output}{Output}
\textbf{Input:} Scene image $I$, object bounding box $B$, network parameters $\psi_{\mathrm{S}}, \psi_{\mathrm{O}}$\\
\textbf{Output:} Predicted paths $\bm{P}=(P_1,P_2,\ldots,P_m)$  \\
\textbf{Scene Analysis} \\
\emph{1. Generate the reward map $\bm{R}_{\mathrm{reward}}$} \\
~~~- Crop out the object image $I_{\mathrm{object}}$ according to $B$, and the scene patches $\bm{q}=\{q_{s_1},q_{s_2},\ldots,q_{s_t}\}$ with an overlapped sliding window on $I$; \\
~~~- \For {$i=1 ~ to ~ t$}{
	~~~ - $\bm{R}_{\mathrm{reward}}(s_i)=\mathcal{F}_{\mathrm{S}} ( I_{\mathrm{object}}, q_{s_i}; \psi_{\mathrm{S}} )$;\\
}
\emph{2. Estimate the object's facing orientation $\theta_{\mathrm{esti}}$}\\
 ~~~- $\theta_{\mathrm{esti}}=\mathcal{F}_{\mathrm{O}} ( I_{\mathrm{object}};\psi_{\mathrm{O}} ) $; \\ 

\textbf{Path Planning}  \\
\emph{1. Find the optimal paths $\bm{P}$}\\
~~~- Obtain the cost map $\bm{R}_{\mathrm{cost}}$ according to Eq. \eqref{rcost}; \\
~~~- Build a directed graph $G$, whose edge weights $W$ are set according to Eq. \eqref{graphweights}; \\
~~~- Compute the shortest paths between $v_{\mathrm{ini}}$ and $\bm{v}_{\mathrm{edge}}$ on graph $G$, and sort them as $\bm{P}=(P_1,P_2,\ldots,P_m)$ based on their lengths $\bm{l}=(l_1,l_2,\ldots,l_m)$ from in an ascending order. \\
\end{algorithm}

In order to solve problem \eqref{optimization} more efficiently and easily, we employ a graph shortest path algorithm. We build a directed graph $G=(V,E)$ whose nodes $v_i$ correspond to the positions $s_i$ of map $\bm{R}_{\mathrm{cost}}$. The weight $W$ of edges $e(v_i,v_j)$ is:
\begin{equation}
\label{graphweights}
W(v_i,v_j)= 
\begin{cases}
\bm{R}_{\mathrm{cost}}(s_j)+ \varepsilon  \mathcal{D}(\theta_{s_j},\theta_{\mathrm{esti}}) & \mbox{for~ (I)},  \\
\bm{R}_{\mathrm{cost}}(s_j) &  \mbox{for~ (II)},\\
+ \infty \quad &  \text{others},
\end{cases}
\end{equation}

\noindent
where
\begin{align}
(I) \text{:} ~~ &  \lVert s_j-s_{\mathrm{ini}} \rVert_1 = d ~ \text{or}~  d\text{+}1,			\notag	\\
& s_{i} ~\text{and}~s_{j} ~ \text{are spatially adjacent}, \notag \\
(II) \text{:} ~  & \lVert s_j-s_{\mathrm{ini}} \rVert_1 \neq d ~ \text{and}~  d\text{+}1, \notag \\
& s_{i} ~\text{and}~s_{j} ~ \text{are spatially adjacent}. 			\notag
\end{align}

\noindent
where $\theta_{s_j}$ is the relative position between $s_{\mathrm{ini}}$ and $s_j$. On graph $G$ we can compute the shortest paths between the node of the initial position $v_{\mathrm{ini}}$ and the nodes of all the edge points $\bm{v}_{\mathrm{edge}}=(v_{\mathrm{edge}_1},v_{\mathrm{edge}_2},\ldots,v_{\mathrm{edge}_m})$ using Dijsktra’s algorithm. These paths are sorted according to their lengths $\bm{l}=(l_1,l_2,\ldots,l_m)$ in an ascending order, represented as $\bm{P}=(P_1,P_2,\ldots,P_m)$. $\bm{P}$ are the top predicted paths for the visual path prediction task. The shortest one $P_1$ is exactly the solution of problem \eqref{optimization} on large scales. The whole path planning procedure is summarized in Algorithm \ref{alg}.

\section{Experiments}
\label{section:experiments}

\subsection{Experimental Setup}
\label{subsection:Baselines, Metric and Datasets}

We give the details on the the network architecture, datasets, the comparison algorithms, and the evaluation metric in the following.

\noindent
{\bf Network Architecture:} 
We build the CNNs based on the popular Caffe toolbox \cite{jia2014caffe}. Fig. \ref{fig:Spatial Matching Network} and Fig. \ref{fig:orientation network} respectively illustrate the network architectures of Spatial Matching Network and Orientation Network. Specifically, in the two figures `conv' represents a convolution layer, `fc' represents a fully connected layer, `pool' represents a max-pooling layer, and `LRN' represents a local response normalization layer. Numbers in the parentheses are respectively kernel size, number of outputs, and stride. All convolutional layers and fully connected layers are followed by ReLU activation function. 

In the experiments, Spatial Matching Network is trained for 2K iterations with a batch size of 256 and learning rate of $10^{-3}$. The input images are uniformly resized to 256$\times$256 and cropped to patches with the size of 227$\times$227. Orientation Network is trained for 10K iterations with a batch size of 256 and learning rate of $10^{-5}$. The input images of Orientation Network are directly resized to 227$\times$227 without any cropping operation, because a part of an object image often cannot represent its exact facing orientation. The weights of models are initialized randomly for a fair comparison with the other algorithms.

\begin{figure}[t]
\begin{center}
   \includegraphics[width=1\linewidth]{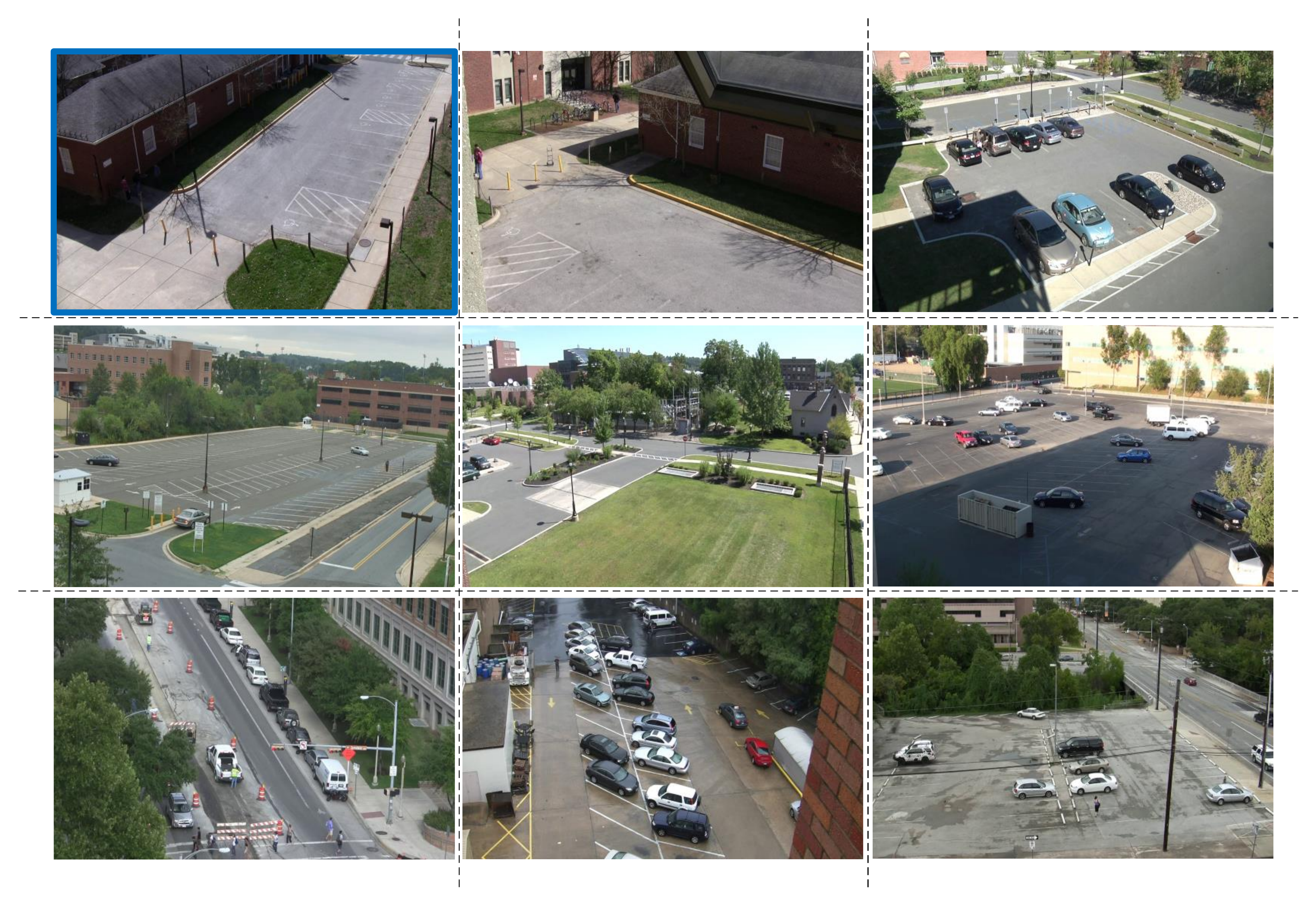}
\end{center}
\caption{The nine scenes of the first evaluation set, which is used in the evaluation of visual path prediction performance. The scenes include different parking lots, streets, and campuses. They are in a semi-birdseye view, and the videos are shot by cameras at different heights and locations to the grounds. The blue box denotes the scene used in the previous work \cite{walker2014patch}. In this paper, we make quantitative experiments on every scene.}
\label{fig:VIRAT}
\end{figure}

\noindent
{\bf Datasets:} For the evaluation of visual path prediction task, we adapt a new large evaluation set. Raw data of this set come from VIRAT Video Dataset Release 2.0 \cite{oh2011large}. VIRAT\footnote{http://www.viratdata.org/} is a public video dataset collected in multiple natural scenes, with people or vehicles performing actions with cluttered backgrounds. It contains a total of 8.5 hours HD videos from 11 different outdoor scenes, with a variety of camera viewpoints, and diverse types of activities which involve both human and vehicles. The ground truth object bounding boxes are manually annotated. Previous work \cite{kitani2012activity,walker2014patch} on path prediction has also built their evaluation set with VIRAT, but with only a single or very few scenes. We do it in a different manner. We select 9 applicable scenes from the total 11 scenes to form our evaluation set. Fig. \ref{fig:VIRAT} shows the chosen scenes clearly, where the coloured box denotes the scene adopted by previous work. Among the total 195 videos, we use 152 videos for training, and 43 videos for testing. From the testing set, we automatically extract objects with at least 200 pixels in length to form a total of 386 testing samples. 

For evaluating the model's generalization capability, we also adapt a novel evaluation set. Raw data of this set come from KIT AIS Dataset\footnote{http://www.ipf.kit.edu/english/code.php}, which comprises aerial image sequences with manually labeled trajectories of the visible vehicles. This dataset is entirely novel to visual path prediction task to our knowledge. It is relatively smaller than VIRAT, so we only use it for testing without training. From the total 9 scenes, we select 8 appropriate scenes and automatically extract 136 samples from the labeled trajectories to construct our evaluation set. The selected trajectories have larger distance between their starting and ending points. 

\noindent
{\bf Comparison methods:} There has been only a little work in the field of visual path
prediction, so in this paper we compare our model with two methods: 
\begin{enumerate}
\item 
\label{method1}
Nearest neighbour searching with SIFT Flow warping \cite{yuen2010data,liu2008sift}. Identical to the implementation in Walker et al. \cite{walker2014patch}, we use a Gist-matching approach \cite{oliva2001modeling} similar to Yuen et al. \cite{yuen2010data}, and warp the labeled path of the nearest neighbour scene into the test scene using SIFT Flow \cite{liu2008sift}.
\item  The mid-level elements based temporal modeling \cite{walker2014patch}. It is the current state-of-the-art approach for visual path prediction task. We use their publicly available implementation code and train a model according to their parameters on the VIRAT dataset. 
\end{enumerate}
In our experiments, all the methods including ours share the same training and testing sets. Because of the larger size of the evaluation set and the higher resolution of the scene images, images are uniformly downsampled into 640$\times$360 for method (1) and (2).

\noindent
{\bf Evaluation metric:} We employ the commonly used \cite{kitani2012activity,walker2014patch} metric: modified Hausdorff distance (MHD) \cite{dubuisson1994modified} as the metric for the distance between two paths. The MHD allows for finding the best local point correspondence and it is robust to outlier points.  Three indicators are used in this paper for comprehensive comparison: (1) top-1, (2) top-5 average and (3) top-10 average. The top-N average means that for a method on a certain testing sample, we first compute the MHDs between the ground truth path and the top-N paths predicted by this method, and then take an average of these distances as the method's performance on this sample.

\subsection{Path Prediction}
\label{subsection:Path Prediction}

\begin{figure*}[t]
\centering
\subfigure{
\begin{minipage}[b]{0.31\linewidth}
\includegraphics[width=1\linewidth]{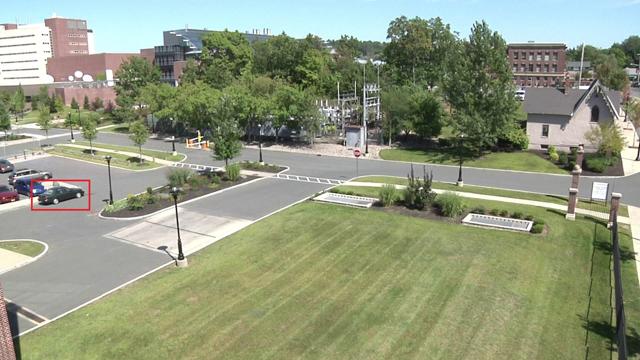}
\end{minipage}
}
\subfigure{
\begin{minipage}[b]{0.31\linewidth}
\includegraphics[width=1\linewidth]{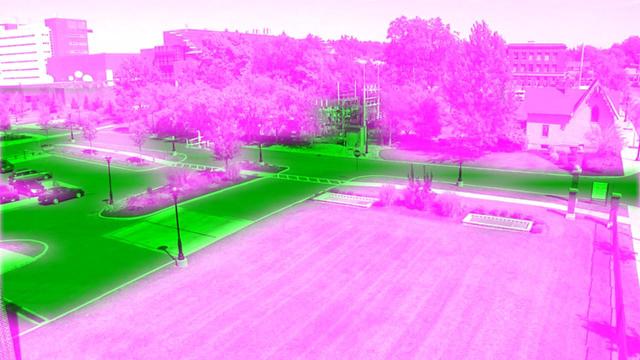}
\end{minipage}
}
\subfigure{
\begin{minipage}[b]{0.31\linewidth}
\includegraphics[width=1\linewidth]{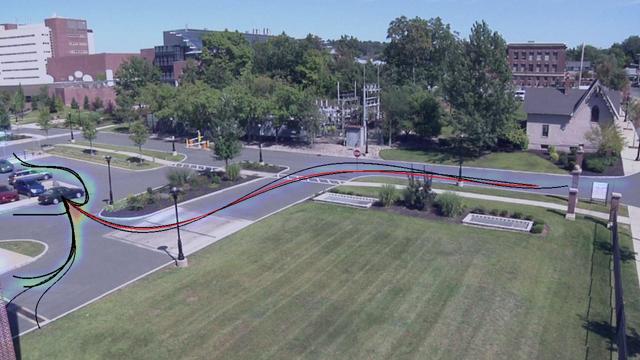}
\end{minipage}
}
\subfigure{
\begin{minipage}[b]{0.31\linewidth}
\includegraphics[width=1\linewidth]{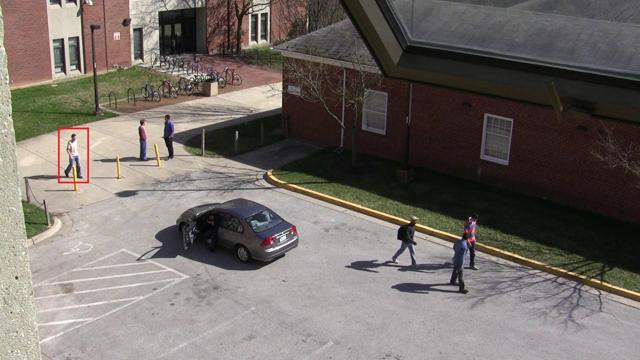}
\end{minipage}
}
\subfigure{
\begin{minipage}[b]{0.31\linewidth}
\includegraphics[width=1\linewidth]{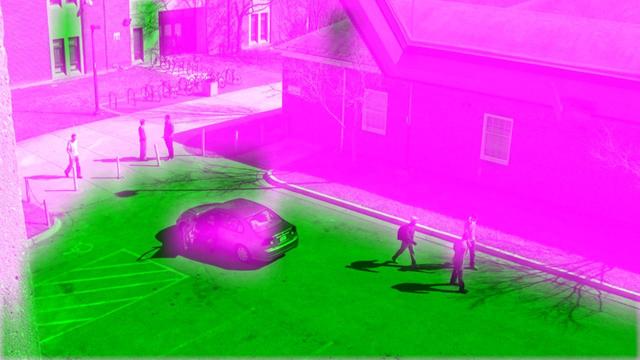}
\end{minipage}
}
\subfigure{
\begin{minipage}[b]{0.31\linewidth}
\includegraphics[width=1\linewidth]{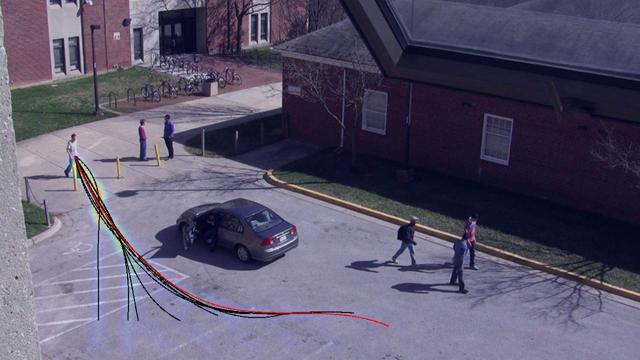}
\end{minipage}
}
\subfigure{
\begin{minipage}[b]{0.31\linewidth}
\includegraphics[width=1\linewidth]{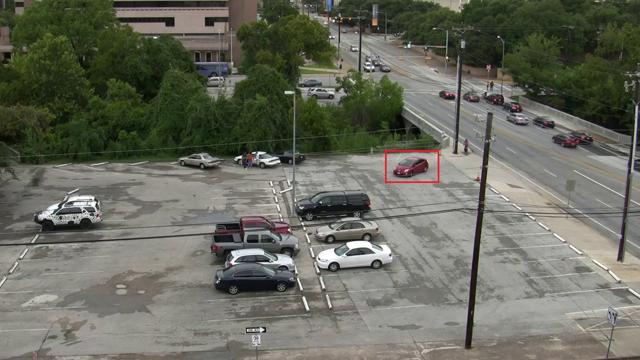}
\end{minipage}
}
\subfigure{
\begin{minipage}[b]{0.31\linewidth}
\includegraphics[width=1\linewidth]{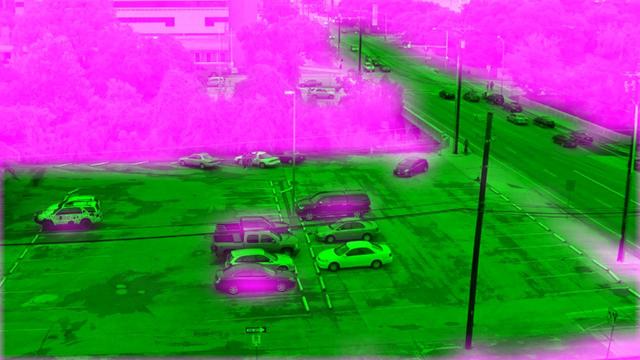}
\end{minipage}
}
\subfigure{
\begin{minipage}[b]{0.31\linewidth}
\includegraphics[width=1\linewidth]{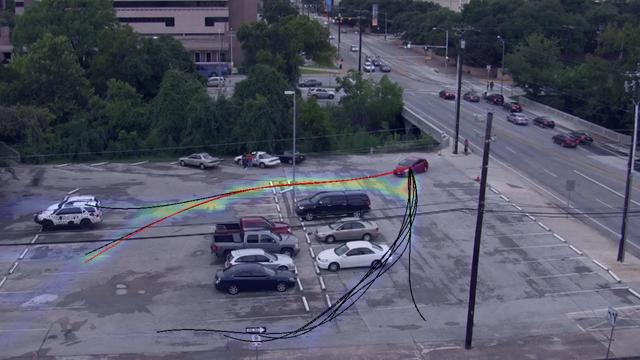}
\end{minipage}
}
\subfigure{
\begin{minipage}[b]{0.31\linewidth}
\includegraphics[width=1\linewidth]{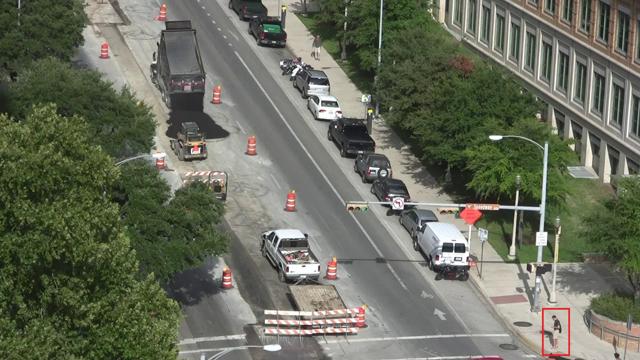}
\end{minipage}
}
\subfigure{
\begin{minipage}[b]{0.31\linewidth}
\includegraphics[width=1\linewidth]{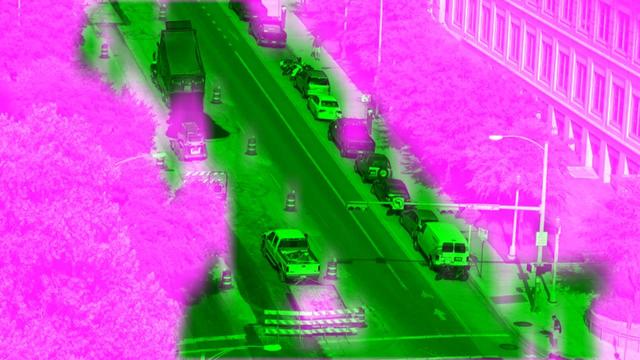}
\end{minipage}
}
\subfigure{
\begin{minipage}[b]{0.31\linewidth}
\includegraphics[width=1\linewidth]{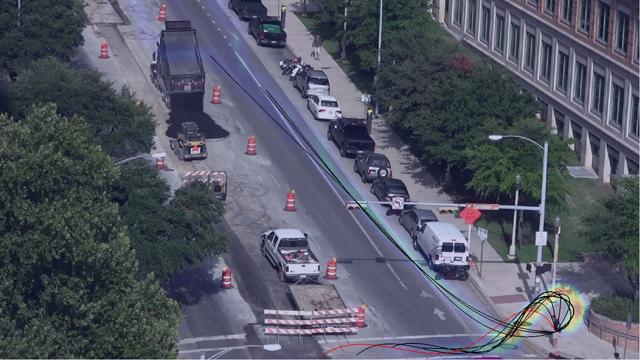}
\end{minipage}
}
\subfigure{
\begin{minipage}[b]{0.31\linewidth}
\includegraphics[width=1\linewidth]{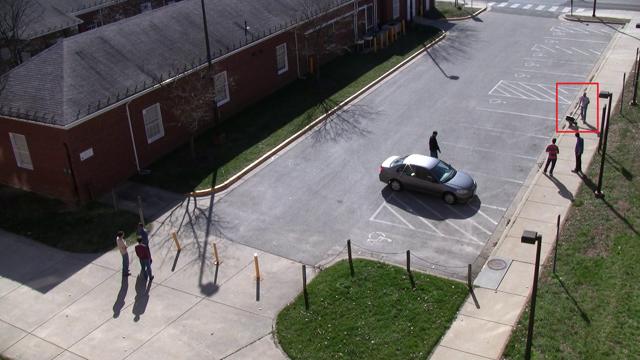}
\centerline{(a) Original Image}
\end{minipage}
}
\subfigure{
\begin{minipage}[b]{0.31\linewidth}
\includegraphics[width=1\linewidth]{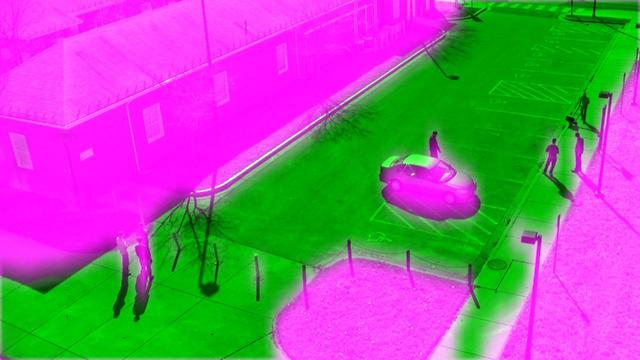}
\centerline{(b) Reward Map}
\end{minipage}
}
\subfigure{
\begin{minipage}[b]{0.31\linewidth}
\includegraphics[width=1\linewidth]{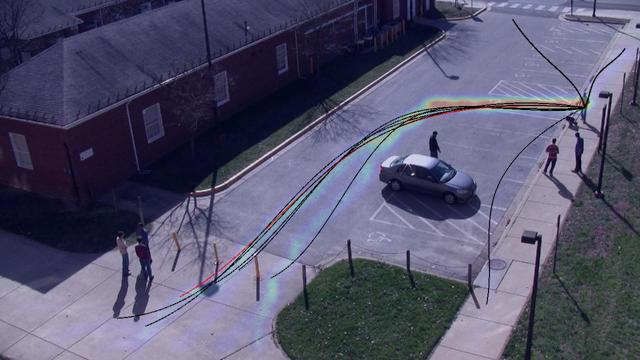}
\centerline{(c) Predicted Paths}
\end{minipage}
}
\caption{Qualitative results generated by our algorithm. Each row represents a sample. The left column shows the input images. Red boxes on them denote the given objects. The middle column shows the generated reward maps. The right column shows predicted top-10 paths. Our framework can output discriminative reward maps and make accurate predictions on a diverse set of scenes.}
\label{fig:qualitative}
\end{figure*}

\begin{figure*}[t]
\centering
\subfigure{
\begin{minipage}[b]{0.23\linewidth}
\includegraphics[width=1\linewidth]{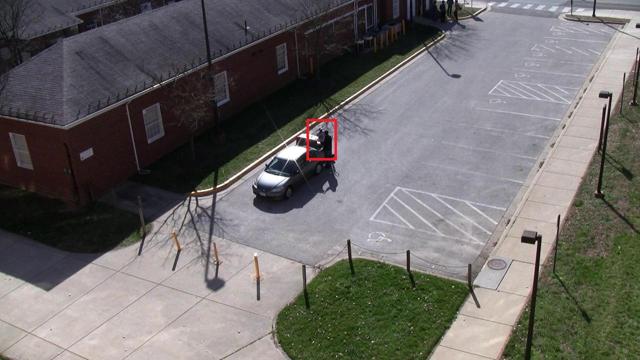}
\end{minipage}
}
\subfigure{
\begin{minipage}[b]{0.23\linewidth}
\includegraphics[width=1\linewidth]{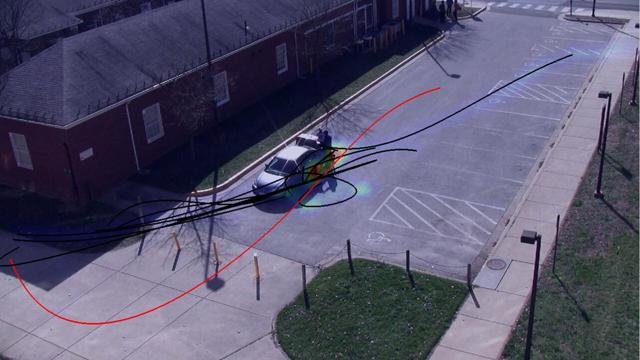}
\end{minipage}
}
\subfigure{
\begin{minipage}[b]{0.23\linewidth}
\includegraphics[width=1\linewidth]{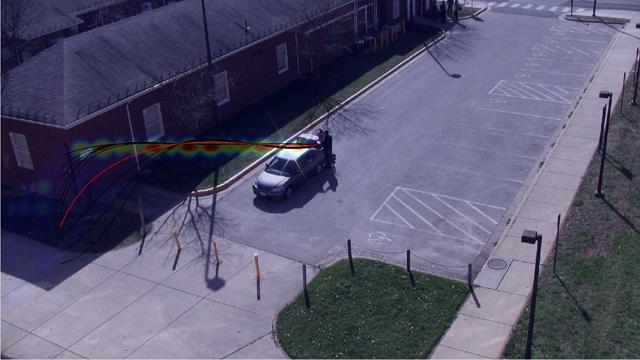}
\end{minipage}
}
\subfigure{
\begin{minipage}[b]{0.23\linewidth}
\includegraphics[width=1\linewidth]{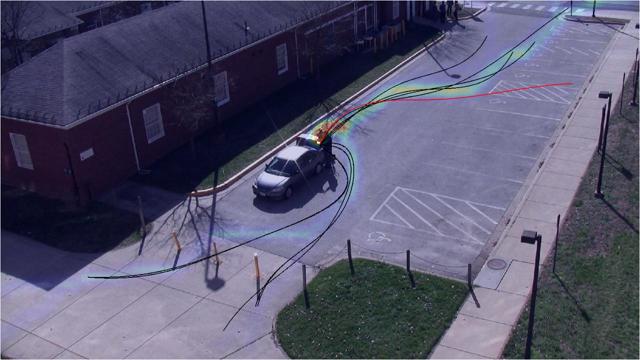}
\end{minipage}
}
\subfigure{
\begin{minipage}[b]{0.23\linewidth}
\includegraphics[width=1\linewidth]{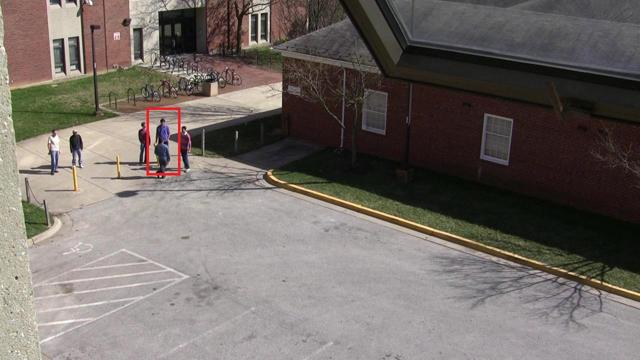}
\end{minipage}
}
\subfigure{
\begin{minipage}[b]{0.23\linewidth}
\includegraphics[width=1\linewidth]{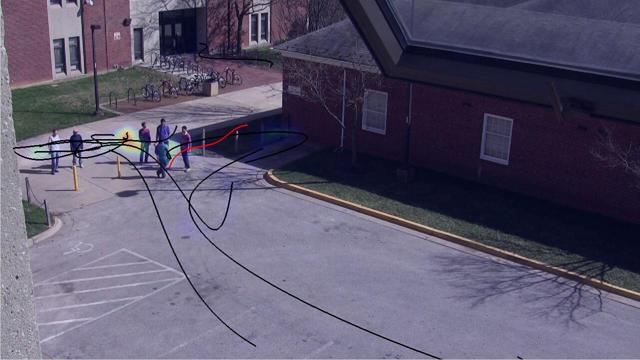}
\end{minipage}
}
\subfigure{
\begin{minipage}[b]{0.23\linewidth}
\includegraphics[width=1\linewidth]{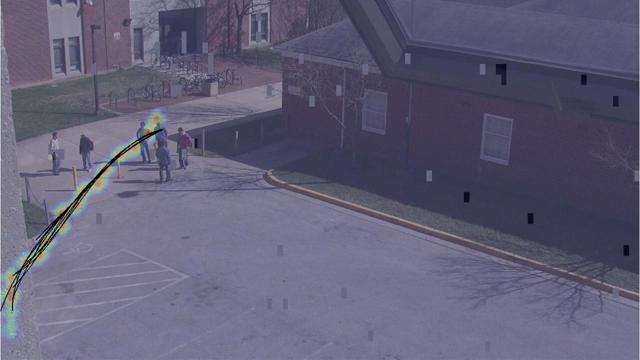}
\end{minipage}
}
\subfigure{
\begin{minipage}[b]{0.23\linewidth}
\includegraphics[width=1\linewidth]{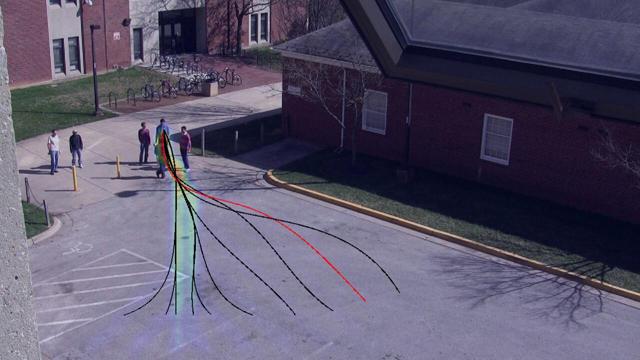}
\end{minipage}
}
\subfigure{
\begin{minipage}[b]{0.23\linewidth}
\includegraphics[width=1\linewidth]{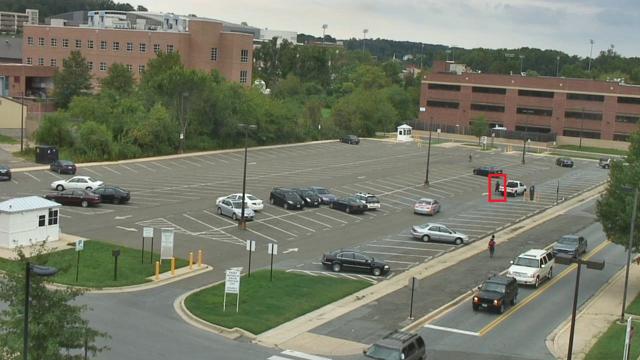}
\end{minipage}
}
\subfigure{
\begin{minipage}[b]{0.23\linewidth}
\includegraphics[width=1\linewidth]{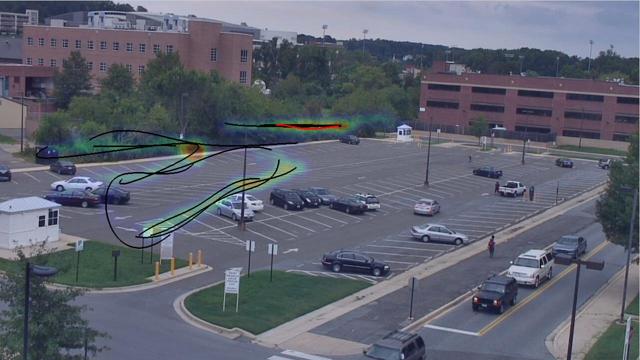}
\end{minipage}
}
\subfigure{
\begin{minipage}[b]{0.23\linewidth}
\includegraphics[width=1\linewidth]{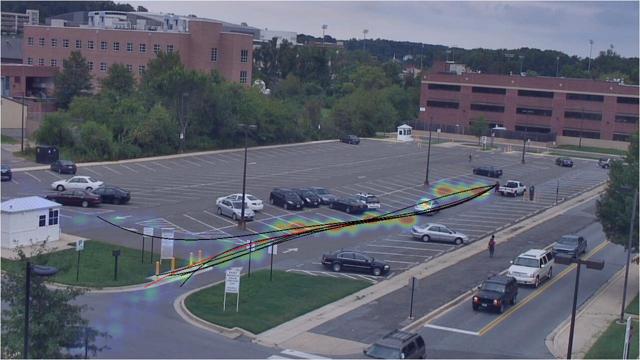}
\end{minipage}
}
\subfigure{
\begin{minipage}[b]{0.23\linewidth}
\includegraphics[width=1\linewidth]{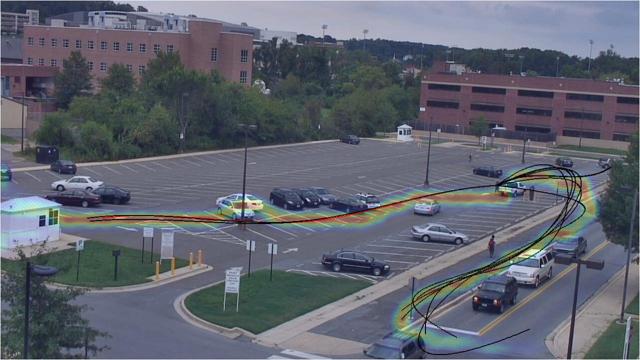}
\end{minipage}
}
\subfigure{
\begin{minipage}[b]{0.23\linewidth}
\includegraphics[width=1\linewidth]{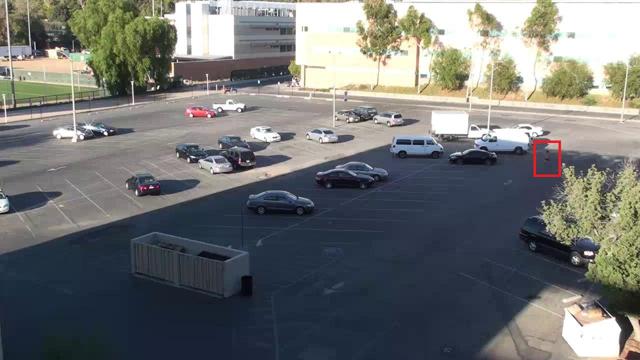}
\end{minipage}
}
\subfigure{
\begin{minipage}[b]{0.23\linewidth}
\includegraphics[width=1\linewidth]{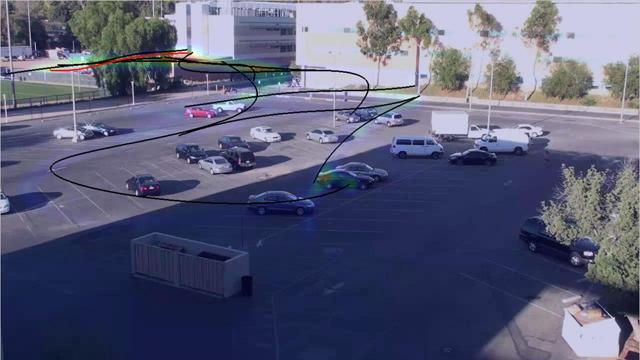}
\end{minipage}
}
\subfigure{
\begin{minipage}[b]{0.23\linewidth}
\includegraphics[width=1\linewidth]{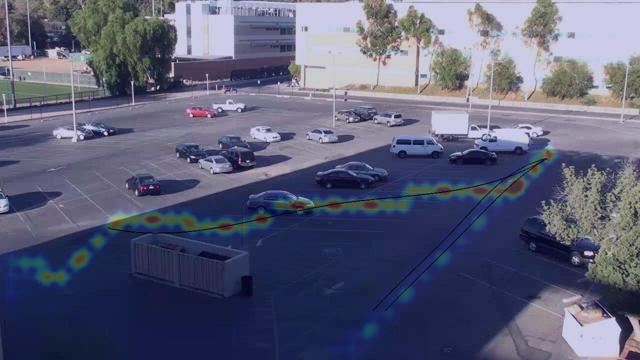}
\end{minipage}
}
\subfigure{
\begin{minipage}[b]{0.23\linewidth}
\includegraphics[width=1\linewidth]{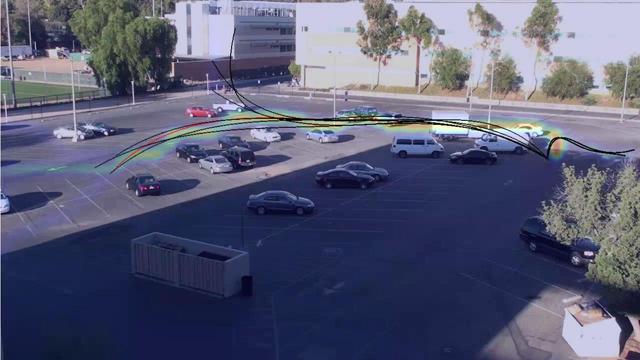}
\end{minipage}
}
\subfigure{
\begin{minipage}[b]{0.23\linewidth}
\includegraphics[width=1\linewidth]{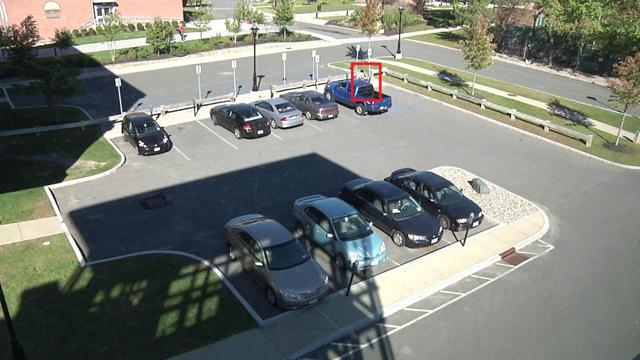}
\end{minipage}
}
\subfigure{
\begin{minipage}[b]{0.23\linewidth}
\includegraphics[width=1\linewidth]{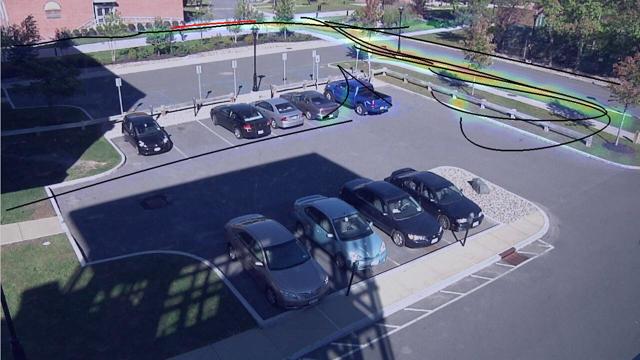}
\end{minipage}
}
\subfigure{
\begin{minipage}[b]{0.23\linewidth}
\includegraphics[width=1\linewidth]{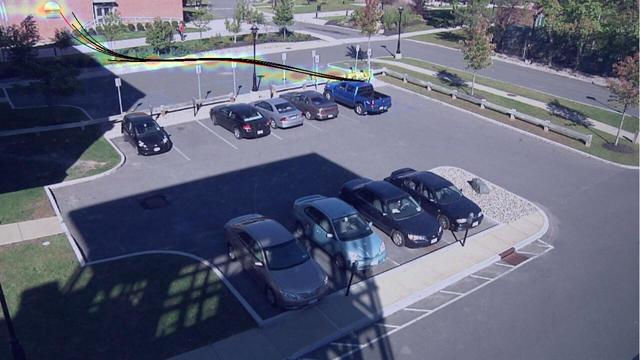}
\end{minipage}
}
\subfigure{
\begin{minipage}[b]{0.23\linewidth}
\includegraphics[width=1\linewidth]{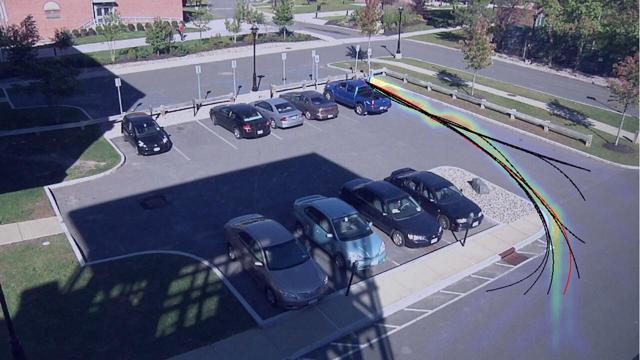}
\end{minipage}
}
\subfigure{
\begin{minipage}[b]{0.23\linewidth}
\includegraphics[width=1\linewidth]{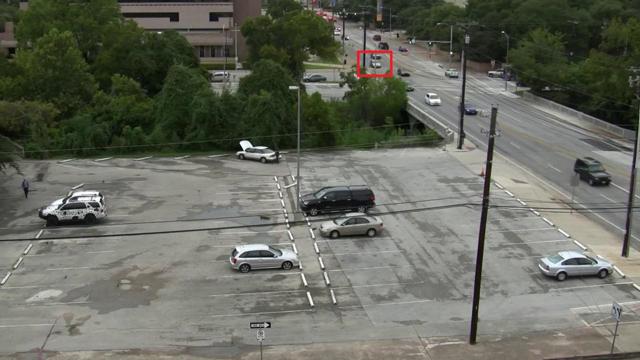}
\centerline{(a) Original Image}
\end{minipage}
}
\subfigure{
\begin{minipage}[b]{0.23\linewidth}
\includegraphics[width=1\linewidth]{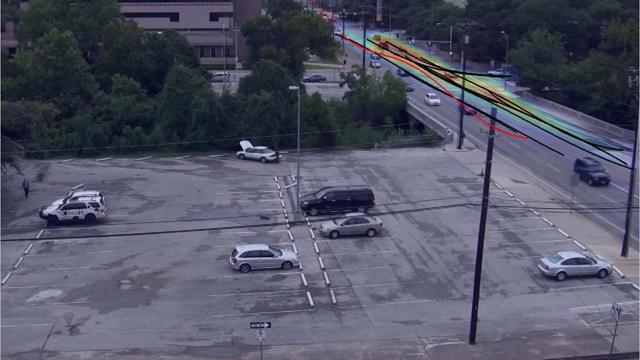}
\centerline{(b) Nearest Neighbour}
\end{minipage}
}
\subfigure{
\begin{minipage}[b]{0.23\linewidth}
\includegraphics[width=1\linewidth]{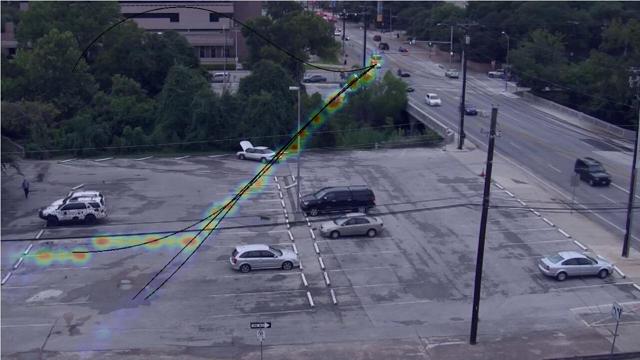}
\centerline{(c) Mid-level Elements }
\end{minipage}
}
\subfigure{
\begin{minipage}[b]{0.23\linewidth}
\includegraphics[width=1\linewidth]{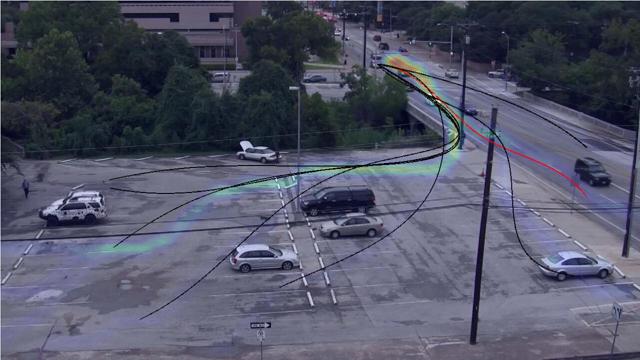}
\centerline{(d) Ours}
\end{minipage}
}
\caption{Some qualitative comparison results. Each row represents a sample. Column (a) shows the input images with red boxes denoting the objects. Column (b), (c) and (d) respectively show the predicted paths generated by different approaches: NN \cite{yuen2010data}, MLE \cite{walker2014patch} and ours. Our approach has better performance in most of the scenes. The predictions generated by our approach are closer to the common sense.}
\label{fig:qualitative2}
\end{figure*}

\noindent
{\bf Qualitative:} Fig. \ref{fig:qualitative} shows some qualitative results generated by our method on different scenes of the evaluation set. Each row represents a sample. The left column is the input images. The middle column shows the reward maps generated by our algorithm, in which those green areas are accessible (high reward) while pink areas are obstacles (low reward). We can see in the maps that the grass, tree and house are detected as low reward, while the road and parking lot are of high reward. Notice the fourth and fifth maps, where the sidewalk is recognized as high reward area for the corresponding pedestrians. The right column shows the predicted paths for corresponding input images, where the red lines represent the top-1 predictions and the black lines represent the other top-10 predictions. Visually, the predicted paths are close to our human's inference. Notice how the predicted paths avoid other objects (cars, pedestrians) or obstacles (grass, trees, buildings) and go along the road. Furthermore, we can see that our framework is able to make correct prediction of the destination. In the third image, the red car will be parked in the square. In the fourth image, the person probably wants to walk across the street. A correct destination estimation will largely improve the performance of path planning.

Besides, we make qualitative comparison among different methods as shown in Fig. \ref{fig:qualitative2}. We select various scenes and objects for testing. Each row represents a testing sample. Column (a) is the input images, in which we mark the given objects with red boxes. Column (b) and (c) show the predicted paths generated by the comparison methods NN \cite{yuen2010data} and MLE \cite{walker2014patch}. Our predictions are shown in column (d). We can see that the NN approach does not give effective performance. It is nearly betting that there have been appropriate paths stored in database. The last image of column (b) shows this clearly where most trajectories of the nearest samples in database are distributed along the road. It is not effective in practical use. MLE approach produces comparatively better performance. However, limited to its visual representation ability on diverse scenes, the predicted paths do not appear reasonable. In the fifth image of column (c), the man would attempt to climb over the fence in front of him. In the sixth image of column (c), the car would attempt to drive across the trees. On most scenes shown in Fig. \ref{fig:qualitative2}, our approach makes reasonable predictions that is consistent with the common sense. Furthermore, our method infers a variety of appropriate optional paths as shown in the first, third and sixth image of column (d).

\begin{table*}[t]
\caption{Quantitative results for visual path prediction task}
\begin{center}
\begin{tabular}{c  c c c c c c c c c | c }
\hlinew{0.8pt} \hline
Scene &  \textbf{A} & \textbf{B} & \textbf{C} & \textbf{D} & \textbf{E} & \textbf{F} & \textbf{G} & \textbf{H} & \textbf{I} & Total\\ 
Samples & 53 & 26 & 21	& 36	& 26 &	41 &	44 &	46	& 93 & 386\\ 

\hline
• & \multicolumn{10}{c}{\textbf{Top-1} }\\ 
\hline
\textbf{NN \cite{yuen2010data}} &  19.57 & 25.47 & 16.15 & 18.19 & 24.78 & 29.16 & 23.16 & 14.84 & 12.31 & 19.12  \\ 

\textbf{MLE \cite{walker2014patch}} & 17.63 &	17.55 &	24.12 &	15.06 &	13.72 &	19.27 &	20.47 &	16.57 &	18.13 	&	17.97 
 \\ 

\textbf{Rewards (ours)} & 20.83 &	20.43 &	\textbf{13.72} &	19.01 &	21.67 &	17.13 &	16.06 &	16.03 &	13.30 &		16.98 
 \\ 

\textbf{Ours}  & \textbf{13.37} &	\textbf{13.81} &	16.34 &	\textbf{13.29} &	\textbf{12.95} &	\textbf{10.99} &	\textbf{10.41} &	\textbf{12.24} &	\textbf{10.42} &	\textbf{12.09}  \\ 

\hline
• & \multicolumn{10}{c}{\textbf{Top-5 Average} }\\ 
\hline
\textbf{NN \cite{yuen2010data}} & 22.34 &	25.75 &	16.35 &	17.10 &	28.89 &	31.09 &	22.86 &	14.65 &	12.72 	&	19.95   \\ 

\textbf{MLE \cite{walker2014patch}} & 17.41 &	17.49 &	22.77 &	15.03 &	13.75 &	19.00 &	20.22 &	16.51 &	18.09 	&	17.78 
 \\ 

\textbf{Rewards (ours)} & 18.87 &	18.80 &	\textbf{13.68} &	17.94 &	20.55 &	17.13 &	17.06 &	13.50 &	11.73 &		15.86 \\ 

\textbf{Ours}  & \textbf{13.21} &	\textbf{13.43} &	15.71 &	\textbf{13.17} &	\textbf{12.78} &	\textbf{11.71} &	\textbf{10.22} &	\textbf{11.60} &	\textbf{10.57} 	&	\textbf{12.00} \\ 
\hline

• & \multicolumn{10}{c}{\textbf{Top-10 Average} }\\ 
\hline
\textbf{NN \cite{yuen2010data}} & 22.81 &	26.31 &	15.86 &	16.19 &	28.31 &	31.38 &	23.37 &	15.88 &	14.42 	&	20.55 
  \\ 

\textbf{MLE \cite{walker2014patch}} & 17.04 &	16.85& 	20.44 &	15.92 &	13.49 	&18.19 	&20.16 &	15.59 &	16.68 	&	17.09 \\ 

\textbf{Rewards (ours)} & 17.62 & 17.43 &\textbf{ 14.97} &	17.57 &	19.65 &	16.16 &	17.27 &	12.24 &	11.16 & 		15.20 \\ 

\textbf{Ours} & \textbf{13.44} &	\textbf{12.89} &	15.59 &	\textbf{12.96} &	\textbf{12.55} &	\textbf{12.11 }&	\textbf{11.79} &	\textbf{11.14} &	\textbf{10.69 }	&	\textbf{12.15} \\ 
\hlinew{0.8pt} \hline
\end{tabular}
\end{center}
\label{table1}
\end{table*}

\begin{table*}[t]
\caption{Quantitative results for generalization capability evaluation}
\begin{center}
\begin{tabular}{c  c c c c c c c c | c }
\hlinew{0.8pt} \hline
Scene No. &  \textbf{1} & \textbf{2} & \textbf{3} & \textbf{4} & \textbf{5} & \textbf{6} & \textbf{7} & \textbf{8}  & Total\\ 
Samples & 5 &	36 &	5 &	6 &	45 &	22 &	6 &	11 & 	136 \\ 

\hline
• & \multicolumn{9}{c}{\textbf{Top-1} }\\ 
\hline
\textbf{NN \cite{yuen2010data}} &\textbf{ 14.54} & 18.75 &	52.77 &	40.67 &	51.58 &	43.38 &	17.67 &	15.54 &		35.35 \\ 

\textbf{MLE \cite{walker2014patch}} &  24.50 &	20.12 &	\textbf{32.20} &	25.66 &	75.45 &	42.70 &	16.19 &	14.05 & 42.26 \\ 

\textbf{Rewards (ours)} & 22.42& 	\textbf{8.57} &	56.96 	&\textbf{23.71}& 	23.97 &	37.52 	&\textbf{15.42} &	10.73 &			21.78  \\
\textbf{Ours}   & 18.23 &	8.99 &	49.24 &	\textbf{23.71} &	\textbf{21.53} &	\textbf{34.19 }&	19.77 &	\textbf{10.14 }	& \textbf{20.25 }
\\
 
\hline
• & \multicolumn{9}{c}{\textbf{Top-5 Average} }\\ 
\hline
\textbf{NN \cite{yuen2010data}} & 17.28 &	21.91 &	50.80 &	34.78 &	64.64 &	44.38 &	16.04 &	16.71 &			40.46 \\ 
\textbf{MLE \cite{walker2014patch}} & 24.28 &	18.76 &	\textbf{32.20} &	25.73 &	75.33 &	42.61 &	16.21 	&14.32 	&	41.87 \\ 

\textbf{Rewards (ours)}& 18.90 	& 6.70& 55.31& 	14.90 &\textbf{	19.46 }	&34.79 &	\textbf{12.16} &	\textbf{8.82} &	18.48 
 \\ 
\textbf{Ours} & \textbf{16.29} &	\textbf{6.63} &	48.76 &\textbf{	12.82} &	19.70 &	\textbf{32.12} &	14.97 &	9.76 	&		\textbf{17.88}  \\ 

\hline
• & \multicolumn{9}{c}{\textbf{Top-10 Average} }\\ 
\hline
\textbf{NN \cite{yuen2010data}} & 17.36 &	20.31 &	53.27 &	34.82 &	61.38 &	46.13 &	19.29 &	15.75 &		39.41 \\ 

\textbf{MLE \cite{walker2014patch}} & 22.92 &	16.22 &	\textbf{43.71} &	25.66 &	75.39 &	42.59 &	16.14 &12.92 		&	41.47 \\ 

\textbf{Rewards (ours)}& 17.68& 	6.56& 	47.84 	&\textbf{13.93} &	\textbf{21.24} &	30.02 	&\textbf{11.78} 	&\textbf{9.65} &17.94\\ 

\textbf{Ours} & \textbf{15.78} &	\textbf{6.37} &	43.82 &	14.80 &	21.47 &	\textbf{27.55} &	12.45 &	10.01 	&		\textbf{17.45}\\
\hlinew{0.8pt} \hline
\end{tabular}
\end{center}
\label{table2}
\end{table*}

\noindent
{\bf Quantitative:} For quantitative evaluation, we compare our method with the competing methods on all scenes in the evaluation set. Table \ref{table1} shows the results on each scene with a total of 386 testing samples. Our method outperforms the comparison methods by large margins on all of the scenes. Compared to the state-of-the-art methods on the entire evaluation set, our method makes 33\%, 33\%, 29\% improvement respectively under the top-1, top-5 average and top-10 average metric. For each scene, the improvement varies from 6\% to 49\% under the top-1 metrics. In addition, the results of our method show a relatively smaller inter-scene variance than the other methods. To some extent, it indicates that our model can be trained and tested robustly on a diverse set of scenes. 

\begin{figure}[t]
\centering
\subfigure{
\begin{minipage}[b]{0.465\linewidth}
\includegraphics[width=1\linewidth]{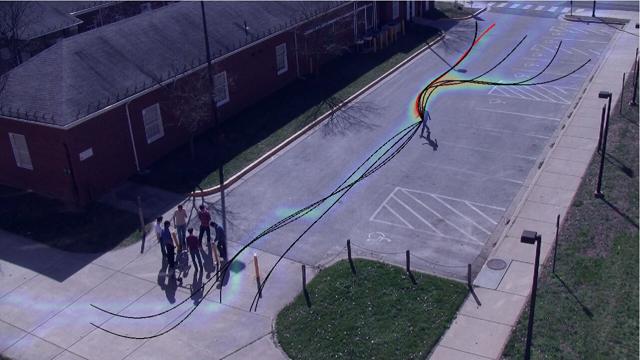}
\end{minipage}
}
\subfigure{
\begin{minipage}[b]{0.465\linewidth}
\includegraphics[width=1\linewidth]{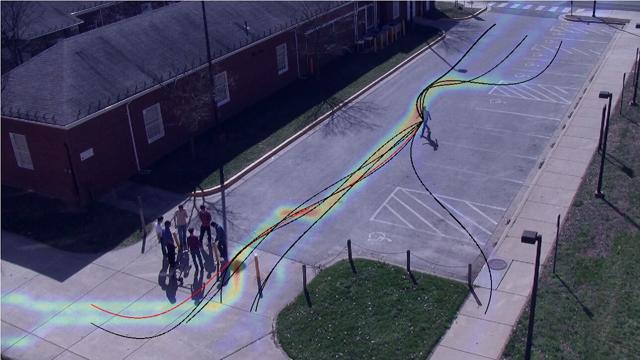}
\end{minipage}
}
\subfigure{
\begin{minipage}[b]{0.465\linewidth}
\includegraphics[width=1\linewidth]{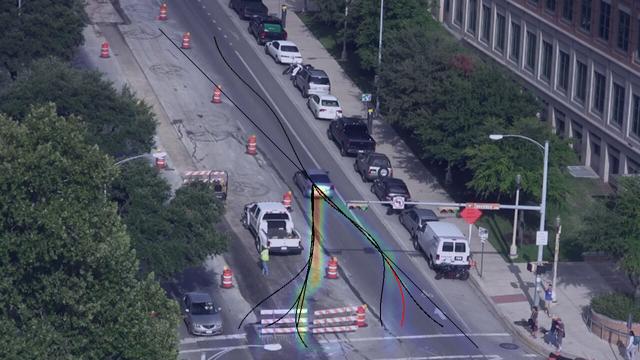}
\end{minipage}
}
\subfigure{
\begin{minipage}[b]{0.465\linewidth}
\includegraphics[width=1\linewidth]{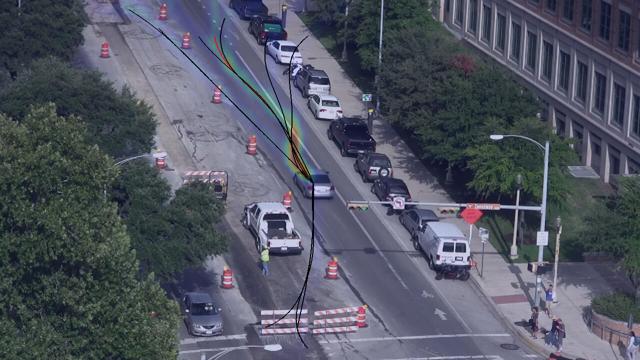}
\end{minipage}
}
\subfigure{
\begin{minipage}[b]{0.465\linewidth}
\includegraphics[width=1\linewidth]{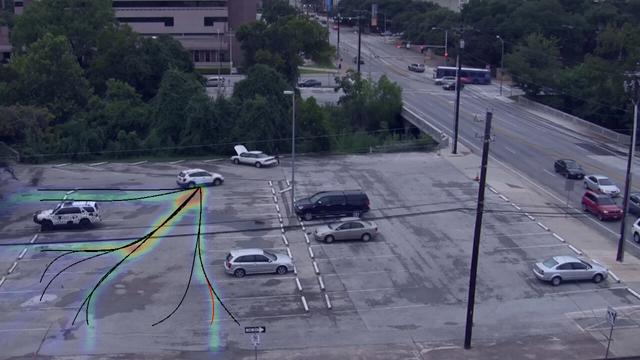}
\centerline{(a) Rewards Only}
\end{minipage}
}
\subfigure{
\begin{minipage}[b]{0.465\linewidth}
\includegraphics[width=1\linewidth]{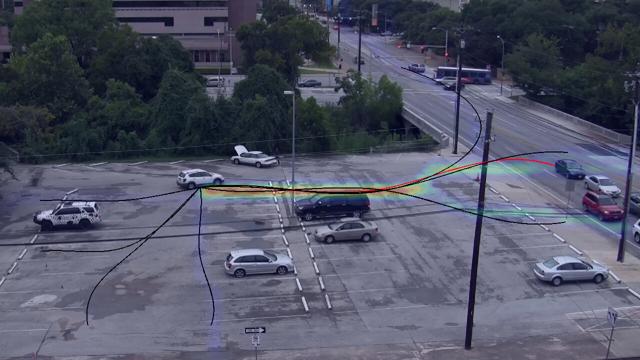}
\centerline{(b) Rewards + Orientation}
\end{minipage}
}
\caption{Comparison between (a) our path planning scheme using only rewards and (b) the complete framework. Orientation Network estimates the facing orientation of the object. With its help, the path planning scheme is able to rectify the paths according to the object's current moving direction, consequently improving the final performance.}
\label{fig:qualitativeours}
\end{figure}

The third row in every sheet shows the results of our rewards only method, for which we only use our reward map for prediction without the help of Orientation Network. It shows 6\%, 11\%, 11\% improvement over the other comparison methods under the three metrics, respectively, demonstrating the value of our Spatial Matching Network. However, the error of the rewards only method is larger than that of our complete framework on most of the scenes. Fig. \ref{fig:qualitativeours} shows a more qualitative comparison between these two methods. It indicates that the temporal context modeled by Orientation Network also offers much help to our complete framework.

\subsection{Generalization Capability}

We have evaluated the visual path prediction performance, where the training set and testing set own the same scenes. However, a robust path prediction framework ought to perform well on novel scenes and objects. In this experiment, we evaluate the generalization capability of the methods on the second evaluation set described in subsection \ref{subsection:Baselines, Metric and Datasets}. We simply test the models on this evaluation set without retraining the models. Parameters of the models remain the same as those in the path prediction experiment of subsection \ref{subsection:Path Prediction}. 

Table \ref{table2} documents the quantitative results of the generalization capability evaluation. Our method respectively makes 43\%, 56\%, 56\% improvement over the comparison methods under the top-1, top-5 average and top-10 average metrics on the entire evaluation set. These improvements are larger than those in the primary experiments as Table \ref{table1}, showing that our method has a better generalization ability than the existing work. Most of the absolute MHD values in Table \ref{table2} have increased, while meantime the inter-scene variance has also increased. This is in line with our intuition that the models have never seen the testing samples in this experiment. In addition, different from the other two methods, the top-5 average metric and top-10 average metric of our method in \ref{table2} show some improvement over the top-1 metric on most of the scenes. To some extent this indicates that our method can explore more proper underlying paths on unknown scenes than the other methods. Compared with the rewards only method, the complete framework performs better on half of the scenes and a little worse on the entire dataset. This indicates that in this experiment the Orientation Network does not help much. It is possibly due to the inadequate training samples.

\section{Conclusion}
In this paper we proposed a deep learning framework to address the visual path prediction problem. The proposed deep learning framework simultaneously performs deep feature learning for visual representation in conjunction with spatio-temporal context modeling, which largely enhances the scene understanding capability. In addition, we presented a unified path planning scheme to infer the future paths on the basis of the analytic results returned by our context models. For comprehensively evaluating the model's performance on the visual path prediction task, we constructed two large benchmark datasets from the adaptation of video tracking datasets. The experimental results demonstrated the effectiveness and robustness of our approach in comparison with the state-of-the-art literature. 

\bibliographystyle{hieeetr}
\bibliography{IEEEabrv,pathprediction}

\end{document}